\documentclass[11pt,a4paper]{article}
\usepackage{times,latexsym}
\usepackage{url}
\usepackage[T1]{fontenc}

\usepackage{graphicx}
\usepackage{subfigure}
\usepackage[font=small,labelfont=bf]{caption}
\usepackage{algorithm}  
\usepackage{algorithmic}
\usepackage{wrapfig}
\usepackage{nicefrac}
\usepackage{booktabs}       
\usepackage{amsfonts}
\usepackage{xcolor}

\usepackage{amsthm,amsmath,bm,xspace}
\usepackage{bbm}
\usepackage{color}
\usepackage{enumitem}
\usepackage{breqn}
\usepackage{pifont}
\usepackage{soul}

\usepackage[acceptedWithA]{tacl2021v1}
\usepackage{tacl2021v1}

\usepackage{xspace,mfirstuc,tabulary}

\newif\iftaclinstructions
\taclinstructionsfalse 
\iftaclinstructions
\renewcommand{\confidential}{}
\renewcommand{\anonsubtext}{(No author info supplied here, for consistency with
TACL-submission anonymization requirements)}
\newcommand{\instr}
\fi

\iftaclpubformat 

\else

\fi

\title{Generate, Annotate, and Learn:
NLP with Synthetic Text}

\author{
 \setlength\tabcolsep{3.5pt}
\begin{tabular}{@{}lllll}
Xuanli He$^1$  &  Islam Nassar$^1$ & Jamie Kiros$^2$ & Gholamreza Haffari$^1$ & Mohammad Norouzi$^2$
\end{tabular}\\
\begin{tabular}{@{}l@{\hspace{1cm}}l}
   $^1$Monash University, Australia & $^2$Google Research, Brain Team
\end{tabular}\\
\texttt{\{xuanli.he1, gholamreza.haffari\}@monash.edu, mnorouzi@google.com}
}

\date{}

\begin{document}
\maketitle
\begin{abstract}
 
 This paper studies the use of language models as a source of synthetic unlabeled  text for NLP.
 We formulate a general framework called ``generate, annotate, and learn (GAL)'' to take advantage of synthetic text within knowledge distillation, self-training, and few-shot learning applications.
 To generate high-quality task-specific text, we either fine-tune LMs on inputs from the task of interest, or prompt large LMs with few examples.
 We use the best available classifier to annotate synthetic text with soft pseudo labels for knowledge distillation and self-training, and use LMs to obtain hard labels for few-shot learning.
 We train new supervised models on the combination of labeled and pseudo-labeled data, which results in significant gains across several applications.
 We investigate key components of GAL and present theoretical and empirical arguments against the use of class-conditional LMs to generate synthetic labeled text instead of unlabeled text.
 GAL achieves new state-of-the-art knowledge distillation results for 6-layer transformers on the GLUE leaderboard. 
\end{abstract}

\def\eg{{\em e.g.,}\xspace}
\def\ie{{\em i.e.,}\xspace}
\def\versus{{\em v.s.}\xspace}
\def\cf{{\em c.f.,}\xspace}

\newcommand{\mohammad}[1]{{\textcolor{green}{ #1}}}
\newcommand{\todo}[1]{{\textcolor{red}{ #1}}}
\newcommand{\xuanli}[1]{{\textcolor{blue}{ #1}}}

\renewcommand{\algorithmicrequire}{\textbf{Input:}}
\renewcommand{\algorithmicensure}{\textbf{Output:}}

\newcommand{\red}[1]{{\bf \textcolor{red}{#1}}}

\newcommand{\gal}{GAL\xspace}

\newcommand{\figleft}{{\em (Left)}}
\newcommand{\figcenter}{{\em (Center)}}
\newcommand{\figright}{{\em (Right)}}
\newcommand{\figtop}{{\em (Top)}}
\newcommand{\figbottom}{{\em (Bottom)}}
\newcommand{\captiona}{{\em (a)}}
\newcommand{\captionb}{{\em (b)}}
\newcommand{\captionc}{{\em (c)}}
\newcommand{\captiond}{{\em (d)}}

\newcommand{\newterm}[1]{{\bf #1}}

\def\figref#1{Figure~\ref{#1}}
\def\Figref#1{Figure~\ref{#1}}
\def\twofigref#1#2{figures \ref{#1} and \ref{#2}}
\def\quadfigref#1#2#3#4{figures \ref{#1}, \ref{#2}, \ref{#3} and \ref{#4}}
\def\tabref#1{Table~\ref{#1}}
\def\Tabref#1{Table~\ref{#1}}
\def\secref#1{section~\ref{#1}}
\def\Secref#1{Section~\ref{#1}}
\def\twosecrefs#1#2{sections \ref{#1} and \ref{#2}}
\def\secrefs#1#2#3{sections \ref{#1}, \ref{#2} and \ref{#3}}
\def\eqref#1{(\ref{#1})}
\def\Eqref#1{Equation~\ref{#1}}
\def\plaineqref#1{\ref{#1}}
\def\chapref#1{chapter~\ref{#1}}
\def\Chapref#1{Chapter~\ref{#1}}
\def\rangechapref#1#2{chapters\ref{#1}--\ref{#2}}
\def\algref#1{algorithm~\ref{#1}}
\def\Algref#1{Algorithm~\ref{#1}}
\def\twoalgref#1#2{Algorithms \ref{#1} and \ref{#2}}
\def\Twoalgref#1#2{Algorithms \ref{#1} and \ref{#2}}
\def\partref#1{part~\ref{#1}}
\def\Partref#1{Part~\ref{#1}}
\def\twopartref#1#2{parts \ref{#1} and \ref{#2}}

\def\ceil#1{\lceil #1 \rceil}
\def\floor#1{\lfloor #1 \rfloor}
\def\1{\bm{1}}
\newcommand{\train}{\mathcal{D}}
\newcommand{\valid}{\mathcal{D_{\mathrm{valid}}}}
\newcommand{\test}{\mathcal{D_{\mathrm{test}}}}

\def\eps{{\epsilon}}

\def\reta{{\textnormal{$\eta$}}}
\def\ra{{\textnormal{a}}}
\def\rb{{\textnormal{b}}}
\def\rc{{\textnormal{c}}}
\def\rd{{\textnormal{d}}}
\def\re{{\textnormal{e}}}
\def\rf{{\textnormal{f}}}
\def\rg{{\textnormal{g}}}
\def\rh{{\textnormal{h}}}
\def\ri{{\textnormal{i}}}
\def\rj{{\textnormal{j}}}
\def\rk{{\textnormal{k}}}
\def\rl{{\textnormal{l}}}
\def\rn{{\textnormal{n}}}
\def\ro{{\textnormal{o}}}
\def\rp{{\textnormal{p}}}
\def\rq{{\textnormal{q}}}
\def\rr{{\textnormal{r}}}
\def\rs{{\textnormal{s}}}
\def\rt{{\textnormal{t}}}
\def\ru{{\textnormal{u}}}
\def\rv{{\textnormal{v}}}
\def\rw{{\textnormal{w}}}
\def\rx{{\textnormal{x}}}
\def\ry{{\textnormal{y}}}
\def\rz{{\textnormal{z}}}

\def\rvepsilon{{\mathbf{\epsilon}}}
\def\rvtheta{{\mathbf{\theta}}}
\def\rva{{\mathbf{a}}}
\def\rvb{{\mathbf{b}}}
\def\rvc{{\mathbf{c}}}
\def\rvd{{\mathbf{d}}}
\def\rve{{\mathbf{e}}}
\def\rvf{{\mathbf{f}}}
\def\rvg{{\mathbf{g}}}
\def\rvh{{\mathbf{h}}}
\def\rvu{{\mathbf{i}}}
\def\rvj{{\mathbf{j}}}
\def\rvk{{\mathbf{k}}}
\def\rvl{{\mathbf{l}}}
\def\rvm{{\mathbf{m}}}
\def\rvn{{\mathbf{n}}}
\def\rvo{{\mathbf{o}}}
\def\rvp{{\mathbf{p}}}
\def\rvq{{\mathbf{q}}}
\def\rvr{{\mathbf{r}}}
\def\rvs{{\mathbf{s}}}
\def\rvt{{\mathbf{t}}}
\def\rvu{{\mathbf{u}}}
\def\rvv{{\mathbf{v}}}
\def\rvw{{\mathbf{w}}}
\def\rvx{{\mathbf{x}}}
\def\rvy{{\mathbf{y}}}
\def\rvz{{\mathbf{z}}}

\def\erva{{\textnormal{a}}}
\def\ervb{{\textnormal{b}}}
\def\ervc{{\textnormal{c}}}
\def\ervd{{\textnormal{d}}}
\def\erve{{\textnormal{e}}}
\def\ervf{{\textnormal{f}}}
\def\ervg{{\textnormal{g}}}
\def\ervh{{\textnormal{h}}}
\def\ervi{{\textnormal{i}}}
\def\ervj{{\textnormal{j}}}
\def\ervk{{\textnormal{k}}}
\def\ervl{{\textnormal{l}}}
\def\ervm{{\textnormal{m}}}
\def\ervn{{\textnormal{n}}}
\def\ervo{{\textnormal{o}}}
\def\ervp{{\textnormal{p}}}
\def\ervq{{\textnormal{q}}}
\def\ervr{{\textnormal{r}}}
\def\ervs{{\textnormal{s}}}
\def\ervt{{\textnormal{t}}}
\def\ervu{{\textnormal{u}}}
\def\ervv{{\textnormal{v}}}
\def\ervw{{\textnormal{w}}}
\def\ervx{{\textnormal{x}}}
\def\ervy{{\textnormal{y}}}
\def\ervz{{\textnormal{z}}}

\def\rmA{{\mathbf{A}}}
\def\rmB{{\mathbf{B}}}
\def\rmC{{\mathbf{C}}}
\def\rmD{{\mathbf{D}}}
\def\rmE{{\mathbf{E}}}
\def\rmF{{\mathbf{F}}}
\def\rmG{{\mathbf{G}}}
\def\rmH{{\mathbf{H}}}
\def\rmI{{\mathbf{I}}}
\def\rmJ{{\mathbf{J}}}
\def\rmK{{\mathbf{K}}}
\def\rmL{{\mathbf{L}}}
\def\rmM{{\mathbf{M}}}
\def\rmN{{\mathbf{N}}}
\def\rmO{{\mathbf{O}}}
\def\rmP{{\mathbf{P}}}
\def\rmQ{{\mathbf{Q}}}
\def\rmR{{\mathbf{R}}}
\def\rmS{{\mathbf{S}}}
\def\rmT{{\mathbf{T}}}
\def\rmU{{\mathbf{U}}}
\def\rmV{{\mathbf{V}}}
\def\rmW{{\mathbf{W}}}
\def\rmX{{\mathbf{X}}}
\def\rmY{{\mathbf{Y}}}
\def\rmZ{{\mathbf{Z}}}

\def\ermA{{\textnormal{A}}}
\def\ermB{{\textnormal{B}}}
\def\ermC{{\textnormal{C}}}
\def\ermD{{\textnormal{D}}}
\def\ermE{{\textnormal{E}}}
\def\ermF{{\textnormal{F}}}
\def\ermG{{\textnormal{G}}}
\def\ermH{{\textnormal{H}}}
\def\ermI{{\textnormal{I}}}
\def\ermJ{{\textnormal{J}}}
\def\ermK{{\textnormal{K}}}
\def\ermL{{\textnormal{L}}}
\def\ermM{{\textnormal{M}}}
\def\ermN{{\textnormal{N}}}
\def\ermO{{\textnormal{O}}}
\def\ermP{{\textnormal{P}}}
\def\ermQ{{\textnormal{Q}}}
\def\ermR{{\textnormal{R}}}
\def\ermS{{\textnormal{S}}}
\def\ermT{{\textnormal{T}}}
\def\ermU{{\textnormal{U}}}
\def\ermV{{\textnormal{V}}}
\def\ermW{{\textnormal{W}}}
\def\ermX{{\textnormal{X}}}
\def\ermY{{\textnormal{Y}}}
\def\ermZ{{\textnormal{Z}}}

\def\vzero{{\bm{0}}}
\def\vone{{\bm{1}}}
\def\vmu{{\bm{\mu}}}
\def\vtheta{{\bm{\theta}}}
\def\va{{\bm{a}}}
\def\vb{{\bm{b}}}
\def\vc{{\bm{c}}}
\def\vd{{\bm{d}}}
\def\ve{{\bm{e}}}
\def\vf{{\bm{f}}}
\def\vg{{\bm{g}}}
\def\vh{{\bm{h}}}
\def\vi{{\bm{i}}}
\def\vj{{\bm{j}}}
\def\vk{{\bm{k}}}
\def\vl{{\bm{l}}}
\def\vm{{\bm{m}}}
\def\vn{{\bm{n}}}
\def\vo{{\bm{o}}}
\def\vp{{\bm{p}}}
\def\vq{{\bm{q}}}
\def\vr{{\bm{r}}}
\def\vs{{\bm{s}}}
\def\vt{{\bm{t}}}
\def\vu{{\bm{u}}}
\def\vv{{\bm{v}}}
\def\vw{{\bm{w}}}
\def\vx{{\bm{x}}}
\def\vtx{\widetilde{\bm{x}}}
\def\vy{{\bm{y}}}
\def\vz{{\bm{z}}}

\def\evalpha{{\alpha}}
\def\evbeta{{\beta}}
\def\evepsilon{{\epsilon}}
\def\evlambda{{\lambda}}
\def\evomega{{\omega}}
\def\evmu{{\mu}}
\def\evpsi{{\psi}}
\def\evsigma{{\sigma}}
\def\evtheta{{\theta}}
\def\eva{{a}}
\def\evb{{b}}
\def\evc{{c}}
\def\evd{{d}}
\def\eve{{e}}
\def\evf{{f}}
\def\evg{{g}}
\def\evh{{h}}
\def\evi{{i}}
\def\evj{{j}}
\def\evk{{k}}
\def\evl{{l}}
\def\evm{{m}}
\def\evn{{n}}
\def\evo{{o}}
\def\evp{{p}}
\def\evq{{q}}
\def\evr{{r}}
\def\evs{{s}}
\def\evt{{t}}
\def\evu{{u}}
\def\evv{{v}}
\def\evw{{w}}
\def\evx{{x}}
\def\evy{{y}}
\def\evz{{z}}

\def\mA{{\bm{A}}}
\def\mB{{\bm{B}}}
\def\mC{{\bm{C}}}
\def\mD{{\bm{D}}}
\def\mE{{\bm{E}}}
\def\mF{{\bm{F}}}
\def\mG{{\bm{G}}}
\def\mH{{\bm{H}}}
\def\mI{{\bm{I}}}
\def\mJ{{\bm{J}}}
\def\mK{{\bm{K}}}
\def\mL{{\bm{L}}}
\def\mM{{\bm{M}}}
\def\mN{{\bm{N}}}
\def\mO{{\bm{O}}}
\def\mP{{\bm{P}}}
\def\mQ{{\bm{Q}}}
\def\mR{{\bm{R}}}
\def\mS{{\bm{S}}}
\def\mT{{\bm{T}}}
\def\mU{{\bm{U}}}
\def\mV{{\bm{V}}}
\def\mW{{\bm{W}}}
\def\mX{{\bm{X}}}
\def\mY{{\bm{Y}}}
\def\mZ{{\bm{Z}}}
\def\mBeta{{\bm{\beta}}}
\def\mPhi{{\bm{\Phi}}}
\def\mLambda{{\bm{\Lambda}}}
\def\mSigma{{\bm{\Sigma}}}

\newcommand{\tens}[1]{\bm{\mathsfit{#1}}}
\def\tA{{\tens{A}}}
\def\tB{{\tens{B}}}
\def\tC{{\tens{C}}}
\def\tD{{\tens{D}}}
\def\tE{{\tens{E}}}
\def\tF{{\tens{F}}}
\def\tG{{\tens{G}}}
\def\tH{{\tens{H}}}
\def\tI{{\tens{I}}}
\def\tJ{{\tens{J}}}
\def\tK{{\tens{K}}}
\def\tL{{\tens{L}}}
\def\tM{{\tens{M}}}
\def\tN{{\tens{N}}}
\def\tO{{\tens{O}}}
\def\tP{{\tens{P}}}
\def\tQ{{\tens{Q}}}
\def\tR{{\tens{R}}}
\def\tS{{\tens{S}}}
\def\tT{{\tens{T}}}
\def\tU{{\tens{U}}}
\def\tV{{\tens{V}}}
\def\tW{{\tens{W}}}
\def\tX{{\tens{X}}}
\def\tY{{\tens{Y}}}
\def\tZ{{\tens{Z}}}

\def\gA{{\mathcal{A}}}
\def\gB{{\mathcal{B}}}
\def\gC{{\mathcal{C}}}
\def\gD{{\mathcal{D}}}
\def\gE{{\mathcal{E}}}
\def\gF{{\mathcal{F}}}
\def\gG{{\mathcal{G}}}
\def\gH{{\mathcal{H}}}
\def\gI{{\mathcal{I}}}
\def\gJ{{\mathcal{J}}}
\def\gK{{\mathcal{K}}}
\def\gL{{\mathcal{L}}}
\def\gM{{\mathcal{M}}}
\def\gN{{\mathcal{N}}}
\def\gO{{\mathcal{O}}}
\def\gP{{\mathcal{P}}}
\def\gQ{{\mathcal{Q}}}
\def\gR{{\mathcal{R}}}
\def\gS{{\mathcal{S}}}
\def\gT{{\mathcal{T}}}
\def\gU{{\mathcal{U}}}
\def\gV{{\mathcal{V}}}
\def\gW{{\mathcal{W}}}
\def\gX{{\mathcal{X}}}
\def\gY{{\mathcal{Y}}}
\def\gZ{{\mathcal{Z}}}

\def\sA{{\mathbb{A}}}
\def\sB{{\mathbb{B}}}
\def\sC{{\mathbb{C}}}
\def\sD{{\mathbb{D}}}
\def\sF{{\mathbb{F}}}
\def\sG{{\mathbb{G}}}
\def\sH{{\mathbb{H}}}
\def\sI{{\mathbb{I}}}
\def\sJ{{\mathbb{J}}}
\def\sK{{\mathbb{K}}}
\def\sL{{\mathbb{L}}}
\def\sM{{\mathbb{M}}}
\def\sN{{\mathbb{N}}}
\def\sO{{\mathbb{O}}}
\def\sP{{\mathbb{P}}}
\def\sQ{{\mathbb{Q}}}
\def\sR{{\mathbb{R}}}
\def\sS{{\mathbb{S}}}
\def\sT{{\mathbb{T}}}
\def\sU{{\mathbb{U}}}
\def\sV{{\mathbb{V}}}
\def\sW{{\mathbb{W}}}
\def\sX{{\mathbb{X}}}
\def\sY{{\mathbb{Y}}}
\def\sZ{{\mathbb{Z}}}

\def\emLambda{{\Lambda}}
\def\emA{{A}}
\def\emB{{B}}
\def\emC{{C}}
\def\emD{{D}}
\def\emE{{E}}
\def\emF{{F}}
\def\emG{{G}}
\def\emH{{H}}
\def\emI{{I}}
\def\emJ{{J}}
\def\emK{{K}}
\def\emL{{L}}
\def\emM{{M}}
\def\emN{{N}}
\def\emO{{O}}
\def\emP{{P}}
\def\emQ{{Q}}
\def\emR{{R}}
\def\emS{{S}}
\def\emT{{T}}
\def\emU{{U}}
\def\emV{{V}}
\def\emW{{W}}
\def\emX{{X}}
\def\emY{{Y}}
\def\emZ{{Z}}
\def\emSigma{{\Sigma}}

\newcommand{\etens}[1]{\mathsfit{#1}}
\def\etLambda{{\etens{\Lambda}}}
\def\etA{{\etens{A}}}
\def\etB{{\etens{B}}}
\def\etC{{\etens{C}}}
\def\etD{{\etens{D}}}
\def\etE{{\etens{E}}}
\def\etF{{\etens{F}}}
\def\etG{{\etens{G}}}
\def\etH{{\etens{H}}}
\def\etI{{\etens{I}}}
\def\etJ{{\etens{J}}}
\def\etK{{\etens{K}}}
\def\etL{{\etens{L}}}
\def\etM{{\etens{M}}}
\def\etN{{\etens{N}}}
\def\etO{{\etens{O}}}
\def\etP{{\etens{P}}}
\def\etQ{{\etens{Q}}}
\def\etR{{\etens{R}}}
\def\etS{{\etens{S}}}
\def\etT{{\etens{T}}}
\def\etU{{\etens{U}}}
\def\etV{{\etens{V}}}
\def\etW{{\etens{W}}}
\def\etX{{\etens{X}}}
\def\etY{{\etens{Y}}}
\def\etZ{{\etens{Z}}}

\newcommand{\pdata}{p_{\rm{data}}}
\newcommand{\ptrain}{\hat{p}_{\rm{data}}}
\newcommand{\Ptrain}{\hat{P}_{\rm{data}}}
\newcommand{\pmodel}{p_{\rm{model}}}
\newcommand{\Pmodel}{P_{\rm{model}}}
\newcommand{\ptildemodel}{\tilde{p}_{\rm{model}}}
\newcommand{\pencode}{p_{\rm{encoder}}}
\newcommand{\pdecode}{p_{\rm{decoder}}}
\newcommand{\precons}{p_{\rm{reconstruct}}}
\newcommand{\one}[1]{\mathbbm{1}{[#1]}}
\newcommand{\laplace}{\mathrm{Laplace}} 

\newcommand{\E}{\mathbb{E}}
\newcommand{\Ls}{\mathcal{L}}
\newcommand{\R}{\mathbb{R}}
\newcommand{\emp}{\tilde{p}}
\newcommand{\lr}{\alpha}
\newcommand{\reg}{\lambda}
\newcommand{\rect}{\mathrm{rectifier}}
\newcommand{\softmax}{\mathrm{softmax}}
\newcommand{\sigmoid}{\sigma}
\newcommand{\softplus}{\zeta}
\newcommand{\KL}{D_{\mathrm{KL}}}
\newcommand{\Var}{\mathrm{Var}}
\newcommand{\standarderror}{\mathrm{SE}}
\newcommand{\Cov}{\mathrm{Cov}}
\newcommand{\normlzero}{L^0}
\newcommand{\normlone}{L^1}
\newcommand{\normltwo}{L^2}
\newcommand{\normlp}{L^p}
\newcommand{\normmax}{L^\infty}

\newcommand{\parents}{Pa} 

\definecolor{myred}{RGB}{215,48,39}
\definecolor{mygreen}{RGB}{26,152,80}
\newcommand{\cmark}{\textcolor{mygreen}{\ding{51}}}
\newcommand{\xmark}{\textcolor{myred}{\ding{55}}}

\vspace{-.3cm}
\section{Introduction}
\vspace{-.2cm}

There is an abundance of unlabeled data in the real world, but task-specific unlabeled data within the scope of a given machine learning problem can be challenging to find.
For instance, one cannot easily find in-domain unlabeled text conforming to the input distribution of a specific Natural Language Processing (NLP) task from the GLUE benchmark~\citep{wang2018glue}.
Some NLP tasks require an input comprising a pair of sentences with a particular relationship between them. 
Moreover, classification datasets typically represent a tailored distribution of data and only include a limited number of class labels.
If task-specific unlabeled data were available, one could adopt self-training~\citep{yarowsky1995unsupervised} to automatically annotate unlabeled data with pseudo labels 
to improve accuracy and robustness of classifiers~\citep{xie2020self, carmon2019unlabeled}. In addition, one can use knowledge distillation~\citep{hinton2015distilling} on fresh
task-specific unlabeled data to more effectively compress deep neural networks and ensembles~\citep{bucilua2006model,chen2020big}.



In the absence of task-specific unlabeled data, one could \emph{retrieve} unlabeled examples from a large and diverse open-domain dataset \citep{du2020self}.
However, such a retrieval-based approach may not scale to problems with complex input schemes, \eg~sentence pairs with certain relations.
%
{Recent work \cite{yang2020g, kumar2020data} has considered the use of Language Models (LMs) like GPT-2~\cite{radford2019language} as a means of data augmentation, showing the effectiveness of this approach for commonsense reasoning and classification tasks. 
%
Existing approaches often consider \emph{class-conditional} generation, where the synthetic data is produced by conditioning on a specified class label. However, it is unclear whether class-conditional generation is best suited for NLP tasks. Furthermore, existing pipelines often make synthetic data generation complicated as one needs to detect and discard low-quality synthetic \emph{labeled} data or optionally re-label data~\cite{yang2020g,vu2021strata}.
For instance, \citet{kumar2020data} observe that it is difficult for sentences generated by label-conditioned GPT-2 to retain the semantics/pragmatics of the conditioning label, leading to poor performance on downstream tasks.
} %
%
%


We unify and simplify existing work on LMs as a data source for NLP and develop 
a general framework called ``generate, annotate, and learn (\gal)''.
The generality of GAL allows us to use LM-generated synthetic data within novel applications such as Knowledge Distillation (KD) and few-shot learning.
GAL builds on recent advances in text generation~\citep{radford2019language,eval-harness} and uses powerful LMs to synthesize
task-specific unlabeled text by fine-tuning or conditioning a large LM on in-distribution examples. We use state-of-the-art classifiers to annotate generated text with soft pseudo labels when possible.
We then combine labeled data and pseudo-labeled data to train more effective supervised models, resulting in significant gains 
on a range of NLP tasks like KD and few-shot learning.

{We present a justification for GAL based on the empirical and vicinal risk minimization frameworks~\citep{vapnik1992principles,chapelle2001vicinal}.
}
{We also investigate key components of GAL. 
We find that even if class-conditional LMs are available for text generation, it is more effective to discard the conditioning labels and let the teacher models produce pseudo labels.
This observation 
is supported by our theoretical and empirical results. 
Accordingly, in contrast to prior work~\cite{yang2020g, vu2021strata}, we advocate for the use of simple unconditional LMs for text synthesis.
Further, we avoid any form of data filtering.
Not surprisingly, we find that the diversity of synthetic text matters. That said, simple unconditional generation given random seeds provides sufficient diversity,
and crafting diverse LM prompts is not needed.}


\noindent{}{In summary:
\begin{itemize}[topsep=0pt, partopsep=0pt, leftmargin=10pt, parsep=0pt, itemsep=1.75pt]
    \item 
    {We develop \gal, a simple and effective approach to the use of LMs for task-specific unlabeled text generation. We show that \gal can used effectively for KD, self-training, and few-shot learning in NLP}. 
    \item We present theoretical and empirical investigations for \gal,
    explaining why it works and why using class-conditional LMs to generate synthetic labeled data is not as effective.
    \item GAL advances KD for NLP and establishes a new SoTA result for a single 6-layer transformer on the GLUE test set.
    It further improves prompt-based few-shot learning, providing an average improvement of 1.3\% on four 4-shot learning NLP tasks, outperforming GPT-3-6B\footnote{Code is available at: \url{https://github.com/xlhex/gal_syntex}}. 
\end{itemize}
}

\vspace{-.2cm}
\section{Related Work}
\vspace{-.2cm}

\noindent \textbf{Data synthesis} 
with large pre-trained language models is closely related to our work~\cite{ kumar2020data, yang2020g, vu2021strata, norouzi2020exemplar}.  
%
\citet{yang2020g} propose a complex scheme, including label-conditioned data generation, data relabeling, data filtering, and two-stage training, to utilize synthetic data. By contrast, we show that a simple mixture of the original data and synthetic unconditionally-generated data can provide sizable gains. Furthermore, we show a  broader use of generative models on KD and few-shot learning. 
\citet{vu2021strata} takes a task augmentation approach and employ conditional generation to produce in-domain synthetic data for an auxiliary language inference (NLI) task, which is then used to initialize the target-task classifier. 
However, not all tasks (\eg grammatical acceptability judgments) can benefit from the NLI-style auxiliary task  ~\cite{wang2019can}.
We aim to directly generate the unlabeled in-domain data for the target task.
Unlike \citet{norouzi2020exemplar}, we do not use instance-based generative models.

More broadly, there has been a recent surge in data synthesis and augmentation in NLP, including rule-based and model-based approaches; see \cite{feng-etal-2021-survey} for a recent survey. 
Data synthesis with grammars has been explored in semantic parsing and natural language understanding, eg see \cite{wang-etal-2015-building,wang2021learning,Marzoev2020UnnaturalLP}.  
Existing approaches to data augmentation for NLP include lexicon replacement, sentence retrieval, and round-trip machine translation~\citep{wang2015s, yu2018qanet, kobayashi-2018-contextual, wu2019conditional, lichtarge2019corpora, wei2019eda, alberti2019synthetic, du2020self, shen2020simple}. We, instead, propose the use of unconditional autoregressive LMs for data augmentation. This is simple, flexible, and powerful.

\noindent \textbf{Self-training} 
is one of the oldest approaches for semi-supervised learning \citep{scudder1965probability, fralick1967learning, 1054472, yarowsky1995unsupervised,eisner-karakos-2005-bootstrapping, ueffing-etal-2007-transductive,du2020self}. 
%
{\citet{abney-2004-understanding,DBLP:conf/uai/HaffariS07} have theoretically analyzed self-training for simple decision lists. 
}  
Recent theoretical work analyzes self-training for linear models, often under the assumption that the data distribution is (nearly) Gaussian~\citep{carmon2019unlabeled,pmlr-v119-raghunathan20a,DBLP:conf/nips/ChenWKM20,pmlr-v119-kumar20c,DBLP:journals/corr/abs-2006-11006}.
\citet{wei2021theoretical} prove that, under ``expansion'' and ``class separation'' assumptions, 
self-training can lead to more accurate neural network classifiers.
We present a theoretical framing of \gal in terms of empirical and vicinal risk minimization \citep{vapnik1992principles, chapelle2001vicinal}.

\noindent \textbf{Knowledge Distillation} (KD)~\cite{bucilua2006model, hinton2015distilling} uses a procedure similar to self-training to distill  knowledge of an expressive teacher model into 
a smaller student model. In contrast, self-distillation~\citep{furlanello2018born, zhang2019your, DBLP:conf/nips/MobahiFB20} uses teacher and student models of equal size, hoping to iteratively refine class labels.
Previous work uses unlabeled data~\citep{bucilua2006model} and adversarial training~\citep{wang2018kdgan} to improve KD. 
We demonstrate that synthetic data generated by unconditional generative models can improve KD on NLP, outperforming strong KD baselines, which often add more complexity and additional hyper-parameters \citep[\eg][]{sun2019patient, jiao2019tinybert, xu2020bert, rashid2021mate}.

\vspace{-.2cm}
\section{Generate, Annotate, and Learn (\gal)}
\label{sec:gal}
\vspace{-.2cm}

Given a labeled dataset $L=\{(\vx_i,y_i)\}_{i=1}^N$, we first train an unconditional domain-specific generative model
$g(\vx)$ on $L_{\vx}=\{\vx_i\}_{i=1}^N$, and then use it to synthesize unlabeled data. 
Such synthetic unlabeled data is used within self-training and KD even in the absence of in-domain unlabeled data.
We restrict our attention to basic KD and self-training methods, even though
GAL can be combined with more sophisticated semi-supervised techniques too.

\begin{figure}[t]
\small
  \centering
 \hspace{-.4cm}~\includegraphics[width=0.5\textwidth]{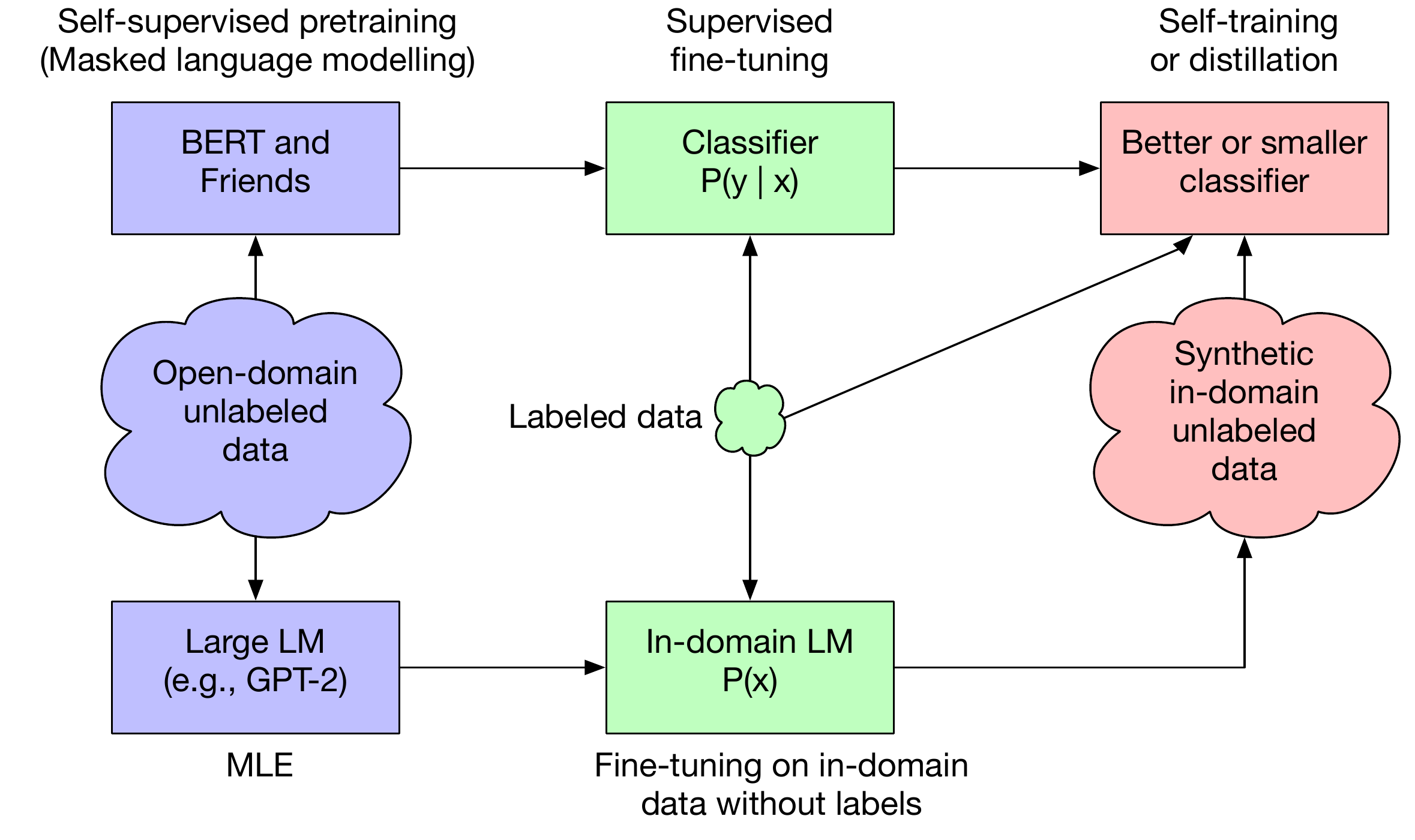}
 \caption{An illustration of \gal for NLP. We use open-domain data once for self-supervised pretraining (\eg~BERT) and once for training a large LM (\eg~GPT-2).
 BERT is fine-tuned on labeled data to yield a classifier for the task of interest. GPT-2 is fine-tuned on the same data without labels to obtain an unconditional task-specific LM, which is used to generate lots of synthetic in-domain unlabeled data for self-training and KD.
 \vspace*{-.6cm}
 }
\label{fig:nlp-gest-summary}
\end{figure}

 The effectiveness of \gal depends on the fidelity and diversity of synthetic examples.
If we had access to the oracle generative process, we were able to obtain the best KD and SSL results, as if we had access to real task-specific unlabeled data.
Our preliminary experiments suggest that large language models are particularly effective within the GAL framework.
%
Hence, as shown in \figref{fig:nlp-gest-summary},
to build the best domain-specific language model, we adopt a large language
model pretrained on lots of open-domain text, and fine-tune it on a given dataset's inputs, \ie~$L_{\vx}$, \textit{ignoring class labels}.
Both our theory and ablations confirm that ignoring class labels is a good idea {(\cf \secref{sec:erm} and \ref{sec:expr})}.
Transferring the knowledge of large language models is particularly beneficial a small input dataset $L_{\vx}$ of text is available~\citep{hernandez2021scaling}.

To improve computational efficiency of \gal,
we do not generate unlabeled data on the fly, but generate as many unconditional samples as possible and store them in a synthetic unlabeled dataset $U$.
We use soft pseudo labels within self-training and KD, {as we empirically found it is more effective than using hard labels on synthetic data}. 


\begin{algorithm}[t]
\begin{algorithmic}[1]
\small
\INPUT Labeled dataset $L=\{(\vx_i,y_i)\}_{i=1}^N$\\
       Initial parameters of a generative model $g_0$\\
       Initial parameters of a classifier $f_0$\\
       A teacher model  $h$\\
\OUTPUT A well-trained student classifier $f_{s}$ after KD \\
 $\triangleright$ unlabeled data generation
\STATE train a generative model $g$ by fine-tuning $g_0$ on $L_{\vx}$ where $L_{\vx} = \{\vx \mid (\vx, y)\in L\}$
\STATE generate ${U} \!=\! \{\vtx_j\}_{j=1}^{kN}$ by drawing $kN$ random samples {\em i.i.d.} from $g(\vx)$ \\ 
 $\triangleright$ knowledge distillation 
\STATE apply $h$ to unlabeled instances of $U$ to get $U'$
\STATE train   $f_s$ by fine-tuning $f_0$ on $L\cup U'$
\STATE \textbf{return} $f_s$
\end{algorithmic}
\caption{$\textrm{\gal-KD}(L$, $g_0$, $f_0$, $h$, $k$)}
\label{alg:gest-kd}
\end{algorithm}

\subsection{Knowledge Distillation with GAL}
KD  distills  knowledge of an expressive teacher model into 
a smaller student model~\citep{hinton2015distilling}.
We pose the following objective function for KD with labeled and synthetic unlabeled data,
\begin{multline}
  \ell_{kd} ~~=~~ \lambda\, \mathbb{E}_{(\vx, y) \sim L}{H}(y, f_{s}(\vx)) +\\
 (1-\lambda)\, \mathbb{E}_{\vtx \sim g(\vx)}  H(h(\vtx), f_{s}(\vtx))
\label{eq:kd:gest}
\end{multline}
where $h$ is the \emph{teacher} model, $f_s$ is the \emph{student} model,  $g$ is the large pre-trained language model (eg GPT2) fine-tuned on the text in the training data $L_{\vx}$.
$H(q, p) = q^\top \log p$ is the softmax cross entropy loss. 
Note the use of $g(\vx)$, approximating the  unknown real data distribution $P(\vx)$ in \eqref{eq:kd:gest}.
%
Algorithm \ref{alg:gest-kd} summarizes the GAL-KD process. 

\subsection{Self-Training with GAL}
Self-training encourages knowledge transfer between a {\em teacher} and a {\em student} model in such a way that the student can outperform the teacher.
\Algref{alg:gest-st} summarizes the \gal-self-training  process. Given the labeled dataset $L$ and the synthetic unlabeled dataset $U$, an initial model denoted $f_1$ is trained using supervised learning on the labeled dataset $L$. Then, at iteration $t$, one adopts $f_t$ as the teacher model to annotate the unlabeled dataset $U$ using {\em pseudo labels}.
In self-training GAL, the student model $f_{t+1}$ is trained to optimize a classification loss on the combination of $L$ and $U$:
%
\begin{multline}
  \ell_{t+1} ~~=~~ \lambda\, \mathbb{E}_{(\vx, y) \sim L} {H}(y, f_{t+1}(\vx)) + \\ (1-\lambda)\, \mathbb{E}_{\vtx \sim g(\vx)}  H(f_t(\vtx), f_{t+1}(\vtx))~.
\label{eq:gest}
\end{multline}
where $\lambda = 0.5$ unless stated otherwise.
Although many different variants of the basic self-training algorithm discussed above exist in the literature, we adopt the simplest variant of self-training and limit hyper-parameter tuning to a bare minimum.

%


\begin{algorithm}[t]
\begin{algorithmic}[1]
\small
\INPUT Labeled dataset $L=\{(\vx_i,y_i)\}_{i=1}^N$\\
       Initial parameters of a generative model $g_0$\\
       Initial parameters of a classifier $f_0$\\
\OUTPUT A better self-training classifier $f_{T+1}$ after $T$ steps \\
 $\triangleright$  unlabeled data generation
\STATE train a generative model $g$ by fine-tuning $g_0$ on $L_{\vx}$ where $L_{\vx} = \{\vx \mid (\vx, y)\in L\}$
\STATE generate ${U} \!=\! \{\vtx_j\}_{j=1}^{kN}$ by drawing $kN$ random samples {\em i.i.d.} from $g(\vx)$ \\ 
 $\triangleright$ self-training
\STATE train a base model $f_1$ by fine-tuning $f_0$ on $L$
\STATE \textbf{for} $t=1$ to $T$ do:
\STATE \quad apply $f_t$ to unlabeled instances of $U$ to get $U'$ 
\STATE \quad train  $f_{t+1}$ by fine-tuning $f_0$ on $L \cup U'$\\
\STATE \textbf{return} $f_{T+1}$
\end{algorithmic}
\caption{$\textrm{\gal-self-training}(L$, $g_0$, $f_0$, $k$, $T)$}
\label{alg:gest-st}
\end{algorithm}


\vspace{-.2cm}
\subsection{{Domain-Specific Text Generation}} 
\vspace{-.1cm}
\label{sec:syn}



{We take a pretrained GPT-2 language model~\citep{radford2019language} and fine-tune it separately on each dataset of interest after removing class labels. We find that training from scratch on these datasets is hopeless, but the larger the pretrained GPT-2 variant, the better the validation perplexity scores are. For tasks modeling a relationship between multiple sentences, we concatenate a separator token ``[SEP]'' between consecutive sentences. {{To alleviate an over-fitting on the training set, we use the best checkpoint evaluated on dev set as our generation engine. Once a fine-tuned GPT-2 model is obtained, we generate new domain-specific data by using top-k random sampling similar to}~\citet{radford2019language}{. We do not feed any prompt to the LM, but a special \textsc{[BOS]} token to initiate the generation chain. A generation episode is terminated when a special \textsc{[EOS]} token is produced. We generate diverse sentences by varying the random seed. After collecting enough synthetic data, we only retain unique sentences. For tasks with $\alpha$ input sentences, we discard generated samples that violate this constraint (approximately 10\% of samples were rejected).}} Finally, we obtain task-specific synthetic data up to $40 \times$ larger than the original training sets. For some samples of generated text for GLUE see \tabref{tab:sst2} and \ref{tab:qqp}. We believe using bigger LMs and larger synthetic datasets will improve our results, but we are constrained by compute resources.
}

\vspace{-.2cm}
\section{An Empirical Risk Minimization Perspective}
\vspace{-.1cm}
\label{sec:erm}

In supervised learning, one seeks to learn a mapping $f$ that given an input $\vx$, predicts a reasonable output $y$.
To define the supervised learning problem formally, one assumes that input-output pairs are drawn from a joint distribution $P$, \ie~$(\vx, y) \sim P(\vx, y)$,
and a loss function $H(y, f(\vx))$ is used to assess the quality of a mapping $f$. This loss is used to define a notion of {\em expected risk}:
\begin{equation}
    R(f) = \mathbb{E}_{P(\vx, y)} H(y, f(\vx)) ~.
\end{equation}

In almost all practical applications $P(\vx, y)$ is unknown. Hence, a labeled dataset of examples $L = \{ (\vx_i, y_i)\}_{i=1}^N$ is used to approximate $R(f)$ as 
\begin{equation}
    \widehat{R}(f) =  \frac{1}{N} \sum\nolimits_{i=1}^N H(y_i, f(\vx_i))~.
\label{eq:rhat}
\end{equation}
This objective function is known as {\em empirical risk}, and learning $f$ through minimizing $\widehat{R}(f)$ is known as the {\em empirical risk minimization} principle~\citep{vapnik1992principles}.
To compensate for the finite sample size in \eqref{eq:rhat}, one typically combines $\widehat{R}(f)$ with a regularizer to improve generalization.

\noindent \textbf{Beyond empirical risk minimization.}
Empirical risk minimization~\eqref{eq:rhat} is motivated as a way to approximate $P(\vx, y)$ through a set 
of Dirac delta functions on labeled examples: $P_{\delta}(\vx, y) = \sum_i \delta(\vx \!=\! \vx_i, y \!=\! y_i) / N$. 
However, this approximation is far from perfect, hence one uses a heldout validation set for early stopping and hyper parameter tuning.

Vicinal risk minimization~\citep{chapelle2001vicinal} approximates expected risk as $\mathbb{E}_{P_{\nu}(\vx, y)} H(y, f(\vx))$, using
a {\em vicinity distribution}, \eg~$\nu(\tilde{\vx}, \tilde{y} \mid \vx, y) = \mathcal{N}(\tilde{\vx} - \vx, \sigma^2)\delta(\tilde{y} = y)$ to
approximate $P(\vx, y)$ as
\begin{equation}
    P_{\nu}(\vx, y) = \frac{1}{N} \sum\nolimits_{i=1}^N \nu(\tilde{\vx}=\vx, \tilde{y}=y \mid \vx_i, y_i)~.
\end{equation}
The goal is to increase the support of each labeled data point and improve the quality and robustness of the risk function.

Recent work on mixup regularization~\citep{zhang2017mixup} proposes an effective way to construct another vicinity distribution by interpolating between two data points and their labels.
Albeit its simplicity, these smoothing techniques tend to improve matters.

\noindent \textbf{Generative models for risk minimization.} 
One can factorize the joint distribution of input-output pairs as $P(\vx, y) = P(\vx) P(y \mid \vx)$.
Accordingly, if one is able to learn a reasonable unconditional generative model of $\vx$ denoted $g(\vx)$, then one can draw a pair $(\vx, y)$ by first drawing $\vx \sim g(\vx)$ and then
using the current instance of $f_t$ to draw $y \sim f_t(\vx)$. Then, one can use $f_t$ and $g$ to approximate expected risk as
\begin{equation}
    R_t(f_{t+1}) = \mathbb{E}_{\vx \sim g(\vx)} \mathbb{E}_{y \sim f_t(\vx)} H(y, f_{t+1}(\vx)) ~.
\label{eq:rt}
\end{equation}
The quality of this approximation highly depends on the quality of $f_t$ and $g$. If $f_t$ is far from an optimal classifier $f^*$ or $g(\vx)$ is far from $P(\vx)$, \eqref{eq:rt} yields a poor approximation.

The expected risk in \eqref{eq:rt} smoothens the risk landscape in complex ways beyond simple Gaussian smoothing and interpolation.
This smoothing is applicable to any continuous, discrete, or structured domain as long as expressive generative models of $P(\vx)$ are available. 
That said, for almost all reasonable loss functions $H$ (\eg~softmax cross entropy and squared error), \eqref{eq:rt} is minimized when $f_{t+1} = f_t$, which
is not ideal, especially when $f_t$ is far from $f^*$.
On the other hand, empirical risk \eqref{eq:rhat} anchors the problem in real labeled examples that are provided as ground truth.

\gal-self-training aims to combine the benefits of \eqref{eq:rhat} and \eqref{eq:rt} via:
\begin{multline}
 R_t(f_{t+1}) ~=~   \frac{\lambda}{N} \sum\nolimits_{i=1}^N H(y_i, f_{t+1}(\vx_i)) +\\ (1-\lambda) \mathbb{E}_{\vx \sim g(\vx)} \mathbb{E}_{y \sim f_t(\vx)} H(y, f_{t+1}(\vx))
\label{eq:gest-risk}
\end{multline}
In this formulation, if $f_t$ represents the minimizer of empirical risk \eqref{eq:rhat}, then $f_{t+1} = f_t$ is the minimizer of \eqref{eq:gest-risk} too.
However, one does not seek the global minimizer of empirical risk, but rather the best performance on heldout data.
If $f_t$ is obtained by stochastic gradient descent on any risk function, but early stopped according to empirical risk on a heldout set,
then using such $f_t$ in \eqref{eq:gest-risk} to define $R_t(f_{t+1})$ promotes the selection of a mapping $f_{t+1}$ that minimizes empirical risk 
while staying close to
the best performing mapping so far (\ie~$f_t$).
This formulation motivates self-training and \gal as regularizers in the functional space
and explains why they can conceivably work.
Although the arguments  are provided here for GAL-self-training,  extending them to GAL-KD is straightforward (omitted due to the space constraint).

\noindent \textbf{How about class-conditional generative models?}
One can also factorize the joint distribution $P(\vx, y)$ as $P(y) P(\vx \mid y)$ and accordingly utilize a
class-conditional generative model $g(\vx \mid y)$ to derive the following expected risk formulation:
\begin{equation}
    R(f) = \mathbb{E}_{y \sim P(y)} \mathbb{E}_{\vx \sim g(\vx|y)} H(y, f_{t+1}(\vx)) ~.
\label{eq:rcc}
\end{equation}
In this setting pseudo labeling is not needed as synthetic data is already labeled.
One can show that the optimal classifier $f_g^*$ that minimizes \eqref{eq:rcc} for the cross-entropy loss is given by,
\begin{equation}
    f_g^*(y \mid \vx) = {g(\vx | y) P(y)} \Big/ {\sum\nolimits_{y'} g(\vx | y') P(y')}~,
\end{equation}
that is turning the class-conditional generative model into a classifier by using the Bayes rule yields the optimal solution.

Provided that the accuracy of generative classifiers on text classification is  behind 
their discriminate counterparts
\citep[\eg][]{ravuri2019classification}, we think substituting \eqref{eq:rcc}
into \eqref{eq:gest-risk} is not a good idea. Essentially, by substituting \eqref{eq:rcc} into the classification objective,
one is regularizing $f$ to remain close to $f_g^*$, which is not an effective strategy if $f_g^*$ is not competitive.
This argument corroborates the evidence from our ablation studies and recent work showing that using class-conditional generative models to
augment supervised learning does not provide big gains~\citep{ravuri2019classification}.

{That said, one can still use class-conditional generative models to synthesize high-fidelity samples. As long as these samples are treated as unlabeled examples and annotated using a classifier, \eg~$f_t$, we believe this is a reasonable approach falling under GAL. Note that our argument above only applies to the scenario that class-conditional generative models are used to synthesize labeled examples. In other words, \gal emphasizes prediction of the labels in the course of the algorithm, rather than having the labels predefined. If one uses the unlabeled synthetic examples from class-conditional generative models, it still aligns to \eqref{eq:gest-risk}, which will be verified in \secref{sec:ablate}.}


\vspace{-.2cm}
\section{Experiments}
\label{sec:expr}
\vspace{-.1cm}

In this section, we assess the effectiveness of \gal on KD, self-training and few-shot learning.


\vspace{-.2cm}
\subsection{{State-of-the-art Results of Knowledge Distillation with \gal on GLUE}}
\vspace{-.1cm}
\label{sec:kd}
We use the GLUE benchmark~\citep{wang2018glue} for our KD experiments; see Appendix~\ref{sec:datasets} for benchmark details. 
%
Our synthetic unlabeled dataset $U$ includes 40$\times$ as many examples as the original
dataset for each task in  GLUE.

It is known that KD on fresh data, unseen during training, performs better~\citep{bucilua2006model, chen2020big} than
KD on original training data. Hence, we investigate the effectiveness of KD using
generated unlabeled data through GAL.

We use the HuggingFace implementation~\citep{wolf-etal-2020-transformers} for KD experiments
and adopt a standard experimental setup consistent with previous work~\citep{sun2019patient, xu2020bert}.
Following \citet{rashid2021mate}, fine-tuned RoBERTa-large (24-layer transformer) represents the teacher
and a DistilRoBERTa (6-layer transformer)~\citep{Sanh2019DistilBERTAD} is used as the student.
We train the student model on $U$ and $L$, where $U$ is annotated by the best RoBERTa-large model, achieving an average score of $86.5$. We then mix $L$ and $U$  with a ratio of 1:4, which is equivalent to $\lambda=0.2$. This ratio works best on the dev set.

\begin{table*}[t]
\setlength\tabcolsep{3pt}
 \caption{\small
    GLUE test results for a 6-layer transformer.
    \gal establishes a new state of the art on KD for NLP.
    Baselines: BERT-Theseus~\citep{xu2020bert}, BERT-PKD~\citep{sun2019patient}, tinyBERT~\citep{jiao2019tinybert} MATE-KD~\citep{rashid2021mate}, DistilRoBERTa~\citep{Sanh2019DistilBERTAD}, and DistilRoBERTa\,+\,KD (standard KD), DistilRoBERTa\,+\,WS (word substitution) and DistilRoBERTa\,+\,RT (round-trip translation). MNLI-m and MNLI-mm indicate matched and mismatched respectively.
    }
    \vspace{-.3cm}
\begin{center}
\scalebox{.92}{
    \begin{tabular}{lccccccccc}
    \toprule
    \textbf{Model} &  \textbf{MNLI(m/mm)} & \textbf{CoLA} & \textbf{SST-2} & \textbf{MRPC} & \textbf{STS-B} & \textbf{QQP} & \textbf{QNLI} & \textbf{RTE}& \textbf{Avg}\\
            							
    \midrule
   \multicolumn{10}{l}{\em Previous work:}\\
     BERT-Theseus & 82.4/82.1 & 47.8 & 92.2 & 87.6/83.2 & 85.6/84.1 & 71.6/89.3 & 89.6 & 66.2 & 78.6 \\
     BERT-PKD & 81.5/81.0 &-& 92.0 & 85.0/79.9 & -  & 70.7/88.9 & 89.0	 & 65.5 & - \\
     tinyBERT & 84.6/83.2 & 51.1 & 93.1 & 87.3/82.6 & 85.0/83.7 & 71.6/89.1 & 90.4 & 70.0	& 79.8 \\
      MATE-KD &  86.2/85.6 & 58.6 & 95.1 & 91.2/88.1 & 88.5/88.4 & 73.0/89.7 & 92.4 & 76.6 & 83.5 \\[.15cm]

   {\em Our results:}\\
     DistilRoBERTa &  83.8/83.4	& 55.9 & 93.2 & 87.4/83.1 & 87.5/87.5 & 71.7/89.1 & 90.6 & 73.3 & 81.2\\
   DistilRoBERTa\,+\,KD & 84.5/84.1 &53.0 & 93.5 & 88.9/85.1 & 88.0/87.4 & 71.9/89.2 & 91.0 & 75.0 & 81.5\\
    {DistilRoBERTa\,+\,WS} & 86.2/85.9 & 52.2 & 94.0 & 89.9/86.4 & 88.7/88.3& 71.7/89.2	 & 91.5 & 76.2 & 82.1 \\
     {DistilRoBERTa\,+\,RT} & 	86.2/85.6 & 55.0 & 94.9 & 90.1/86.5	& 89.2/88.9	& 72.5/89.7	& 92.1 & 77.2 & 82.9 \\
    DistilRoBERTa\,+\,\gal& 86.9/86.4 &58.6 & 95.3 & 91.6/88.7	& 89.9/89.5	& 73.0/89.9	&  92.7 & 79.7	& 84.3\\


   
     \bottomrule
    \end{tabular}
    \vspace{-.15cm}
    }
    \vspace{-.4cm}
    \label{tab:kd}
\end{center}
\end{table*}

\tabref{tab:kd} shows the results of individual 6-layer transformers on the GLUE test set. All of the baselines use an identical student architecture.
\gal achieves the best entry on the GLUE leaderboard, marking a new state-of-the-art for KD on NLP. It outperforms strong KD baselines such as
DistilRoBERTa~\citep{Sanh2019DistilBERTAD}, BERT-PKD~\citep{sun2019patient}, BERT-Theseus~\citep{xu2020bert}, tinyBERT~\citep{jiao2019tinybert} and MATE-KD~\citep{rashid2021mate}.
It also outperforms our own DistilRoBERTa+KD baseline, which learns from soft labels produced by an identical RoBERTa-large ensemble on the original labeled dataset.
While the use of soft labels outperform the vanilla fine-tuned DistilRoBERTa model, it significantly underperforms our KD+\gal baseline. {{We also compare with two strong data-augmentation baselines, round-trip translation (RT)}~\cite{yu2018qanet, shleifer2019low} {and word substitutions (WS)}~\cite{jiao2019tinybert,wei2019eda} {For RT, We generate 40$\times$ unlabeled data using German as the bridge language (English \textrightarrow German\textrightarrow English). The translations are generated via the best model in WMT19}~\citep{ng2019facebook}{. We use the codebase from}~\citet{jiao2019tinybert}{ to conduct WS data augmentation. We mirror the KD experimental setup of GAL for both RT and WS. Although DistilRoBERTa+RT and DistilRoBERTa+WS are better than vanilla DistilRoBERTa and KD variants, they still drastically underperforms our approach.}}



\begin{table*}[t]
\setlength\tabcolsep{2pt} 
 \caption{RoBERTa base and \gal self-training results on GLUE dev sets, averaged across 5 independent runs (numbers in the subscript indicate the error bar, \ie standard deviation divided by $\sqrt{5}$.).}
    \vspace{-.3cm}
    \centering
  
\scalebox{.95}{

    \begin{tabular}{lccccccccc}
    \toprule
            \textbf{Model} &  \textbf{MNLI} & \textbf{CoLA} & \textbf{SST-2} & \textbf{MRPC} & \textbf{STS-B} & \textbf{QQP} & \textbf{QNLI} & \textbf{RTE} & \textbf{Avg}\\
    \midrule 
    RoBERTa base &87.7 $_{\mathrm{0.1}}$ & 63.6 $_{\mathrm{0.4}}$& 94.8 $_{\mathrm{0.1}}$& 90.1 $_{\mathrm{0.4}}$& 90.8 $_{\mathrm{0.1}}$& 91.5 $_{\mathrm{0.1}}$& 92.6 $_{\mathrm{0.1}}$& 78.8	$_{\mathrm{0.4}}$&86.2 \\
    \quad +\,\gal (iter 1) & 87.9 $_{\mathrm{0.1}}$ & 65.1 $_{\mathrm{0.5}}$& 95.3 $_{\mathrm{0.1}}$& 91.7 $_{\mathrm{0.5}}$& 91.4 $_{\mathrm{0.1}}$& 91.8 $_{\mathrm{0.1}}$& 93.1 $_{\mathrm{0.1}}$& 81.4 $_{\mathrm{0.4}}$&87.2\\
    \quad +\,\gal (iter 2) &88.0 $_{\mathrm{0.1}}$& 65.2 $_{\mathrm{0.5}}$& 95.3 $_{\mathrm{0.1}}$& 92.2 $_{\mathrm{0.4}}$& 91.5 $_{\mathrm{0.1}}$& 91.7 $_{\mathrm{0.1}}$& 93.2 $_{\mathrm{0.1}}$& 82.4 $_{\mathrm{0.5}}$&{\bf 87.4}\\
    \quad +\,\gal (iter 3)& 87.9 $_{\mathrm{0.1}}$ & 65.5 $_{\mathrm{0.5}}$& 95.3 $_{\mathrm{0.1}}$ & 92.2 $_{\mathrm{0.5}}$ & 91.7 $_{\mathrm{0.2}}$& 91.7 $_{\mathrm{0.1}}$& 93.2 $_{\mathrm{0.1}}$& 82.0	$_{\mathrm{0.5}}$&{\bf 87.4}\\
    \midrule
    RoBERTa base + self-distillation \quad & 88.1 $_{\mathrm{0.1}}$ & 63.7 $_{\mathrm{0.5}}$ & 95.2 $_{\mathrm{0.1}}$ & 90.3 $_{\mathrm{0.4}}$ & 90.4 $_{\mathrm{0.1}}$ & 91.5 $_{\mathrm{0.1}}$ & 93.1 $_{\mathrm{0.1}}$ & 79.7 $_{\mathrm{0.5}}$ & 86.5\\
     \bottomrule
    \end{tabular}
    }
     \vspace{-.4cm}
    \label{tab:self_train}
\end{table*}

\vspace{-.2cm}
\subsection{{Self-Training with \gal on GLUE}}
\label{sec:nlp}
\vspace{-.1cm}

We fine-tune pretrained RoBERTa model provided by fairseq~\citep{ott2019fairseq} on each GLUE task.
Fine-tuned RoBERTa serves as the first teacher model for self-training.
Each student model is initialized with the original pretrained RoBERTa and fine-tuned with exactly the same hyper-parameters as suggested by fairseq~\citep{ott2019fairseq}.
We combine the labeled dataset $L$ and the synthetic dataset $U$ with a ratio of 1:1, by oversampling labeled data. 
This corresponds to $\lambda = 0.5$ in Eq.~\eqref{eq:gest-risk}.

\tabref{tab:self_train} shows that \gal provides an average improvement of +1.3\% over RoBERTa-base. 
We see consistent improvements with more \gal iterations, but performance saturates after three iterations.
We further compare our approach with a self-distillation~\citep{furlanello2018born} baseline, in which the teacher and student models use the same architecture and transfer knowledge via the original labeled training set.
Although self-distillation provides a slight improvement, the gains from \gal are more significant.



We delve deeper and combine \gal self-training with RoBERTa-large and report test results for both single model and ensemble model in \tabref{tab:g_test}. We observe consistent gains coming from \gal on RoBERTa-large. Our results underperform the latest and biggest LMs from the GLUE leaderboard, but we are optimistic that GAL can be effectively combined with enormous LMs to provide additional gains.

In addition to the GLUE benchmark, Appendix~\ref{app:ic-task} shows the applicability of \gal to two image classification tasks as a proof of concept, but more advanced techniques such as Mixup~\citep{zhang2017mixup} are needed to bridge the gap with the state-of-the-art.

\begin{table*}[h]
\setlength\tabcolsep{4pt} 
 \caption{\small RoBERTa-large with \gal self-training and SoTA methods evaluated on GLUE test sets. The benefit of GAL on single models is larger than ensembles. It appears that self-training reduce the variance of models.
 Baselines including much larger models: RoBERTa-large \citep{liu2019roberta}, ELECTRA~\citep{Clark2020ELECTRA}, T5~\citep{raffel2020exploring}, ERNIE~\citep{sun2019ernie}, and DeBERTa~\citep{he2020deberta}. MNLI-m and MNLI-mm indicate matched and mismatched respectively. 
    }
\begin{center}
\scalebox{.88}{
    \begin{tabular}{lccccccccc}
    \toprule
            \textbf{Model} &  \textbf{MNLI(m/mm)} & \textbf{CoLA} & \textbf{SST-2} & \textbf{MRPC} & \textbf{STS-B} & \textbf{QQP} & \textbf{QNLI} & \textbf{RTE}& \textbf{Avg}\\
            \midrule
        \multicolumn{10}{l}{\em Individual Models (our implementation):}\\
    RoBERTa-large  & 90.1/89.7 &63.8&96.1 & 91.2/88.3 & 90.9/90.7 & 72.5/89.6 & 94.5 & 85.9 & 86.5\\
 RoBERTa-large\,+\,\gal &90.2/89.8 & 66.2 & 96.4 & 92.0/89.2 & 90.7/90.5 & 73.6/89.9 & 95.0 &86.3 &87.1 \\[.15cm]
     \multicolumn{10}{l}{\em Ensemble Models (our implementation):}\\
RoBERTa-large &91.2/90.5 & 66.8 & 96.9 & 92.8/90.3 & 91.9/91.6 & 74.5/90.4 & 95.5 & 87.7 & 87.9 \\
RoBERTa-large\,+\,\gal &91.0/90.7 & 67.9 & 97.1 & 93.1/90.8 & 91.6/91.4 & 74.5/90.4 & 95.8 & 88.2& 88.2 \\[.15cm]
{\em State-of-the-art:}\\
RoBERTa-large &90.8/90.2 & 67.8 & 96.7 & 92.3/89.8 & 92.2/91.9 & 74.3/90.3 & 95.4 & 88.2 & 88.0 \\
ELECTRA & 91.3/90.8 & 71.7 & 97.1 & 93.1/90.7 & 92.9/92.5 & 75.6/90.8 & 95.8 & 89.8	 & 89.2 \\
T5 & 92.2/91.9 & 71.6 & 97.5 & 92.8/90.4 & 93.1/92.8 & 75.1/90.6 & 96.9 & 92.8 & 89.8 \\
ERNIE	& 91.9/91.4 & 74.4 & 97.8 & 93.9/91.8 & 93.0/92.6 & 75.2/90.9 & 97.3 & 92.0 & 90.2\\
DeBERTa	& 91.9/91.6 & 71.5 & 97.5 & 94.0/92.0 & 92.9/92.6 & 76.2/90.8 & 99.2 & 93.2 & 90.3 \\
     \bottomrule
    \end{tabular}
    }
    \end{center}
 
    \vspace{-.2cm}
    \label{tab:g_test}
\end{table*}

\subsection{Prompt-based Few-shot Experiments}
GPT3~\citep{brown2020language} has introduced an optimization-free paradigm for few-shot learning for NLP. Without updating the parameters, large LMs can correctly predict the labels of the inputs by conditioning on a prompt, which consists of an instruction, a few labeled instances and a new unlabeled input.
We apply \gal to prompt-based few-shot learning. Specifically, we present $k$ labeled examples as a prompt to GPT-J~\citep{gpt-j}, an open-sourced re-implementation of GPT-3-6B, and generate $m$ synthetic examples, followed by the corresponding labels. Note that to mitigate noisy outputs, the generation of each synthetic example only conditions on the original $k$ labeled examples. Finally, we concatenate the original $k$ examples and $m$ synthetic examples, and conduct a $(k+m)$-shot learning experiment with GPT-J.

\begin{table*}[h]
 \centering
 \caption{Few-shot learning results for GPT-J (6B)~\citep{gpt-j} on four NLP datasets. Accuracy is reported for these datasets. 
    }
    \begin{tabular}{lccccc}
    \toprule
            \textbf{Model} &  \textbf{SST-2} &\textbf{PIQA} & \textbf{COPA} & \textbf{BoolQ} & \textbf{Avg}\\
            \midrule
            4-shot & 89.8 $_{\mathrm{0.8}}$ & 76.0 $_{\mathrm{1.4}}$ & 79.0 $_{\mathrm{1.5}}$	& 64.3 $_{\mathrm{0.8}}$ & 77.3\\
            8-shot & 91.3 $_{\mathrm{0.8}}$ & 76.2 $_{\mathrm{1.2}}$ & 79.0 $_{\mathrm{1.5}}$& 66.2 $_{\mathrm{0.8}}$ & 78.2\\
            16-shot &  92.7 $_{\mathrm{0.6}}$ & 77.0 $_{\mathrm{0.9}}$ & 81.0 $_{\mathrm{1.1}}$& 66.8 $_{\mathrm{0.8}}$ & 79.4\\
            \midrule
    
            4-shot +\,\ synthetic 12-shot (\gal) & 91.5 $_{\mathrm{0.7}}$& 76.7 $_{\mathrm{1.0}}$ &  80.0 $_{\mathrm{1.2}}$& 65.9 $_{\mathrm{0.8}}$ & 78.5\\
    \bottomrule
    
    \end{tabular}
    \vspace{-.3cm}
    \label{tab:gptj}
\end{table*}

\citet{brown2020language} studied a total of 51 few-shot learning tasks. Studying all of these tasks is prohibitively expensive. Thus, we filter tasks by following these two steps. First, since generating $m$ synthetic examples for each test instance is computationally expensive, we exclude tasks that have more than 5k test examples. Second, we filter tasks on which GPT-3-6B achieves a score lower than 65\% (please refer to Table H.1 in \citet{brown2020language} for more details). After applying the filtering steps, we use four datasets: SST-2~\citep{wang2018glue}, PIQA~\citep{bisk2020piqa}, COPA and BoolQ~\citep{wang2019superglue} as the testbed. We notice that in order to generate valid synthetic data, GPT-J  requires to see at least 4 labeled examples. In addition, at most 16 examples of BoolQ can be fed into GPT-J without truncation. Thus, we set $k$ and $m$ to 4 and 12 respectively. As seen in \tabref{tab:gptj}, GAL leads to an average improvement of 1.2\% over 4-shot learning, and reduces the gap between 4-shot and 16-shot learning.  We noticed that the quality of some generated examples is low. We believe the performance of few-shot learning can be further improved with high-quality instances. One solution is to generate  many synthetic examples, and  select a high-quality subset. Since each test instance conditions on distinct labeled instances, one has to generate different synthetic instances for each test example from GPT-J, which causes expensive computation. Due to such computational constraints, we leave the investigation of data selection strategies  to the future work.

\begin{table}[h]
\caption{\gal with various GPT-2 model sizes on GLUE dev sets. NA indicates a  RoBERTa base model. We \textbf{bold} the best numbers.}
    \centering
  \scalebox{1}{
    \begin{tabular}{lcccc}
    \toprule
            \textbf{GPT-2} & \textbf{SST-2} &  \textbf{RTE} & \textbf{MRPC} & \textbf{CoLA}\\
            \midrule
            NA &  94.8 & 78.8 & 90.1 & 63.6 \\
            small &  \textbf{95.5}  & 81.3  & 90.9  & 63.9 \\
            medium &  95.3  & 81.3  & 91.3 & 63.7 \\
            large &  95.3  & \textbf{81.4}  & \textbf{91.7}   & \textbf{65.1} \\
    \bottomrule
    \end{tabular}
}
    \label{tab:self_gpt}
\end{table}

\subsection{Ablating Components of GAL on GLUE}
\label{sec:ablate}

\begin{table}[h]
\small
\caption{\gal with soft \versus hard pseudo labels on GLUE dev sets. We \textbf{bold} the best numbers.}
    \centering
    \begin{tabular}{lcccc}
    \toprule
            \textbf{Pseudo label} & \textbf{SST-2} &  \textbf{RTE} & \textbf{MRPC} & \textbf{CoLA}\\
            \midrule
            hard &  95.0 &80.7 &90.8 &63.0 \\
            soft &  \textbf{95.3} & \textbf{81.4} & \textbf{91.7} & \textbf{65.1} \\
    \bottomrule    
    \end{tabular}
    \label{tab:self_logits}
    \vspace{-.4cm}
\end{table}

\begin{table*}[h!]
\small
\caption{Synthetic data from class-conditional LMs underperforms \gal and RoBERTa on GLUE dev sets.}
    \centering
    \begin{tabular}{lccccc}
    \toprule
             \textbf{Generative model} & \textbf{Labeled synthetic data} &\textbf{SST-2} &  \textbf{RTE} & \textbf{MRPC} & \textbf{CoLA}\\
             \midrule
             None (baseline) & - &94.8 & 78.8 & 90.1 & 63.6 \\
            \midrule
            Class-conditional LM   &  \cmark & 92.9 & 74.4 & 86.0 & 58.4\\
            Unconditional LM (\gal) & \xmark & 95.3 & 81.4 & 91.7 & 65.1 \\
            Class-conditional LM (GAL) & \xmark & 95.4 & 81.0  & 91.4 &  65.2 \\
    \bottomrule    
    \end{tabular}
    \vspace{-.3cm}
    \label{tab:self_label}
\end{table*}

We conduct an in-depth study of different components of \gal on GLUE datasets. Unless stated otherwise, we use a RoBERTa-base model with a combination of the original training data and $40\!\times$ synthetic data for each self-training experiment.

\noindent \textbf{GPT-2 model size.} \citet{radford2019language} present a few variants of the GPT-2 model including \textit{GPT-2}, \textit{GPT-2-medium}, \textit{GPT-2-large},
and \textit{GPT-2-XL}. Larger GPT-2 models yield better perplexity scores and higher generation quality. We utilize these models except GPT-2-XL within the \gal framework to study the impact of the generative model's quality on downstream task's performance. \tabref{tab:self_gpt} shows that regardless of the GPT-2 model sizes, \gal consistently surpasses the vanilla RoBERTa base. Moreover, SST-2 and RTE datasets are not sensitive to the capacity of  GPT-2,
but higher quality synthetic text improves the results on MRPC and CoLA datasets. We leave investigation of GPT-2-XL and even larger LMs such as GPT-3~\citep{brown2020language} to future work.

\noindent\textbf{Soft v.s. hard pseudo label.} We investigate the use of soft and hard
pseudo labels within the \gal framework.
The results in \tabref{tab:self_logits} suggest that \gal using soft pseudo labels is more effective than hard labels on the GLUE benchmark.
This finding is compatible with the intuition that soft labels enable
measuring the functional similarity of neural networks better~\citep{hinton2015distilling}.

\noindent \textbf{Class-conditional synthetic data generation.}
Previous work~\citep{kumar2020data, ravuri2019classification} suggests that it is challenging 
to utilize labeled synthetic data from class-conditional generative models to boost the accuracy of text and image classifiers. 
Our theory in Section~\ref{sec:erm} points to the potential drawback of class-conditional synthetic data.
We empirically study this phenomenon, by fine-tuning GPT-2 in a class-conditional manner. Then we utilize its synthetic examples in two different cases: 1) labeled synthetic examples and 2) unlabeled synthetic examples.
\tabref{tab:self_label} shows that not only class-conditional LMs underperform unconditional LMs in our \gal framework, 
but also they are much worse than the baseline, when using the pre-defined labels. Nevertheless, if we apply \gal to these examples, the class-conditional LM is on-par with the unconditional one, which corroborates the importance of the annotation step in \gal. We provide more analysis in Appendix \ref{app:pseudo_label}.

\section{Limitations}
{This work demonstrates that one can leverage synthetic in-domain data generated by powerful pre-trained generative models. For simplicity, we do not employ any filtering avenue to retain diverse but high-quality data points. However, previous works have shown that advanced filtering approaches can further improve the performance}~\cite{sohn2020fixmatch, du2020self, yang2020g}. {Given that the improvements in the self-training are not sizeable, we believe it is worth imposing filtering methods on the synthetic data to mitigate the side effects caused by the noisy data points.}

{Although we examine the effectiveness of GAL on various classification tasks, we still focus on the sentence-level tasks. Because of the superior performance on sentence-level tasks, there has been a surge of interest shift to document-level tasks, such as document-level machine translation}~\citep{miculicich-etal-2018-document,voita2018context, maruf2018document},  {and document summarization}~\citep{rush2015neural, nallapati2016abstractive}, {\em{etc.}} {As these tasks suffer from data scarcity, one can leverage GAL to synthesize more data points. However, previous works have shown that GPT-2 has difficulty generating coherent text requiring long-range dependency}~\cite{orbach-goldberg-2020-facts2story,guan2020knowledge}. {Thus, such a limitation may hinder the application of GAL to document-level tasks.}

{In addition, the label space of the studied tasks is not as complex as the structured prediction tasks, such as machine translation, dialog system, question answering, {\em{etc.}}} {However, we believe one can smoothly adapt GAL to these tasks as well. Let us consider machine translation (MT), as a canonical structured prediction task. 
Prior works have shown that one can use (real) monolingual data, in either source or the target language, through data augmentation}~\citep{sennrich-etal-2016-improving} {or knowledge distillation}~\citep{kim2016sequence}{ to improve the structured prediction tasks. This suggests a promising avenue for future research on using synthetically generate monolingual data to improve MT for specialized domains where even monolingual data is scarce.} 

{Furthermore,}  \citet{vu2021generalised}{ suggests that one can leverage a retrieval-based approach to obtain monolingual sentences from the generic data stores. This retrieved monolingual  data is then employed to improve the translation quality in a domain adaptation setting.
This suggests that a GAL-based approach to synthetically generate monolingual text is a promising method to improve MT for specialized domains -- an interesting direction for future research. }

\section{Conclusion}
\label{sec:concl}


We present Generate, Annotate, and Learn (\gal): a framework for self-training and knowledge distillation with generated unlabeled data. 
We motivate \gal from an expected risk minimization perspective and demonstrate both theoretically and empirically that the use of unconditional generative models
for synthetic data generation is more effective than class-conditional generative models, previously used in the literature.
\gal leverages advances in large pretrained language models to help supervised learning and can have implications for 
learning from limited labeled data. 
\gal significantly helps improve knowledge distillation and prompt-based few-shot learning. In addition, a concurrent work~\citep{gowal2021improving} has shown that using generated images can enhance the robustness of images classifiers. We will explore this direction on NLP tasks in the future. Finally, We hope that \gal will stimulate new research on the evaluation and development of large language models.



\section*{Acknowledgments}
We would like to thank the anonymous reviewers and action editor André F.T. Martins for their comments and suggestions on this work. The computational resources of this work are partly supported by the Multi-modal Australian ScienceS Imaging and Visualisation Environment (MASSIVE) (\url{www.massive.org.au}). This material is partly based on research sponsored by Air Force Research Laboratory and DARPA under agreement number FA8750-19-2-0501. The U.S. Government is authorized to reproduce and distribute reprints for Governmental purposes notwithstanding any copyright notation thereon.

\bibliography{refs}
\bibliographystyle{acl_natbib}

\iftaclpubformat

\onecolumn

\appendix
\section{GAL on Text Classification Tasks}

\subsection{Datasets}
\label{sec:datasets}
The statistics of GLUE  are reported in \tabref{tab:data}.

\begin{table*}[h!]
\caption{Summary of GLUE tasks used for evaluation of \gal. STS-B is a regression task, so \#classes is not applicable.}
\centering
\scalebox{0.9}{
    \begin{tabular}{lllrrrc}
    \toprule
            \textbf{Dataset} & \textbf{task} & \textbf{domain} & \textbf{\#train} & \textbf{\#dev} & \textbf{\#test} & \textbf{\#classes}\\
    \midrule  
    \quad SST-2 & sentiment analysis & movie reviews & 67k & 872 & 1.8k & 2\\
    \quad QQP & paraphrase &social QA questions & 364k & 40k & 391k &2 \\
    \quad QNLI & QA/natural language inference & Wikipedia & 105k & 5k & 5.4k & 2\\
    \quad RTE & natural language inference & news, Wikipedia & 2.5k & 277 & 3k& 2 \\
    \quad MNLI & natural language inference &misc. & 393k & 20k & 20k & 3 \\
    \quad MRPC &  paraphrase& news & 3.7k & 408 & 1.7k &2\\
    \quad CoLA & acceptability & misc. & 8.5k & 1043& 1k & 2\\
    \quad STS-B & sentence similarity &misc. & 5.8k & 15k & 1.4k & $-$ \\
     \bottomrule
    \end{tabular}
    }
    \vspace*{-.1cm}
    \label{tab:data}
\end{table*}

\subsection{GPT-2 for classification}
\label{app:gpt2}
{We have conducted additional experiments, where we fine-tune GPT-2 as a classifier. We have considered two variants of the GPT-2 model. The first varant is the original GPT-2 model (GPT2-original) pre-trained on open-domain text. The second variant is the GPT-2 model that was fine-tuned on the inputs of each task separately (GPT-2-finetuned). This model was used to generate task-specific (synthetic) unlabeled data. Finally, we also consider self-training with GAL on top of GPT2-original. Specifically, we use the GPT-2-finetuned model to synthesize 40x in-domain unlabeled data. Then we apply self-training to GPT-2-original, where the data is a combination of the original labeled data and pseudo-labeled synthetic data}. \tabref{tab:gpt2-gal} {suggests that the gains of GAL come from the pseudo-labeled synthetic data, {{\em i.e.,}} both synthetic unlabeled data and teacher's knowledge. Without the generation of synthetic unlabeled data, the domain-specific knowledge embedded in GPT-2-finetuned model cannot be utilized. As such, GPT-2-finetuned model is inferior to the GPT2-original model. Since RoBERTa-large is superior to GPT-2 models, RoBERTa-large+GAL also significantly outperform the GPT-2 counterpart.}


\begin{table*}[h]
 \caption{GLUE test results of using GPT-2 and RoBERTa-large as classification models. }
\begin{center}
\scalebox{.85}{
    \begin{tabular}{lccccccccc}
    \toprule
    \textbf{Model} &  \textbf{MNLI} & \textbf{CoLA} & \textbf{SST-2} & \textbf{MRPC} & \textbf{STS-B} & \textbf{QQP} & \textbf{QNLI} & \textbf{RTE}& \textbf{Avg}\\
            							
    \midrule
  

     GPT-2-original & 85.9/85.6 & 54.8 & 94.5 & 86.9/82.2 & 86.3/85.2 & 72.5/89.3  & 91.2	& 69.8 & 80.9 \\
     GPT-2-finetuned & 85.8/85.5 & 40.9	& 94.5 & 87.0/81.0 & 85.6/84.3 & 71.4/88.5 & 	91.5 &	69.0	& 78.8 \\
    GPT-2-original+\gal & 86.2/85.8 & 55.7 & 94.7 & 87.9/83.4 & 86.9/85.9 & 72.6/89.4	 & 91.9 & 70.6 & 81.5\\
    \midrule
    RoBERTa-large  & 90.1/89.7 &63.8&96.1 & 91.2/88.3 & 90.9/90.7 & 72.5/89.6 & 94.5 & 85.9 & 86.5\\
   RoBERTa-large\,+\,\gal &90.2/89.8 & 66.2 & 96.4 & 92.0/89.2 & 90.7/90.5 & 73.6/89.9 & 95.0 &86.3 &87.1 \\
     \bottomrule
    \end{tabular}
    \vspace{-.3cm}
    }
    \label{tab:gpt2-gal}
\end{center}
\end{table*}

\begin{table}[h]
\caption{Performance of GPT2 annotation, RoBERTa annotation and conditioning labels on 100 random examples from the synthetic RTE dataset generated by a class-conditional LM.}
\vspace{-.2cm}
    \centering
    \scalebox{0.82}{
    \begin{tabular}{l@{\hspace{.25cm}}c@{\hspace{.25cm}}c@{\hspace{.25cm}}c@{\hspace{.25cm}}c}
    \toprule
             \textbf{Label type} & \textbf{Accuracy} &  \textbf{F1} & \textbf{Precision} & \textbf{Recall}\\
             \midrule
             GPT2 & 86.0 & 87.0 & 88.7 & 85.5 \\
             RoBERTa  & 90.0  & 91.4  &  100.0 &  84.1 \\
     conditioning label & 72.0  & 71.4  &  66.0  & 77.8 \\ 

    \bottomrule    
    \end{tabular}
    }
    \vspace{-.4cm}
    \label{tab:human_eval}
\end{table}

\subsection{Importance of Pseudo-labels}
\label{app:pseudo_label}
We have argued and demonstrated that using class-conditional generative models to generate \textit{labeled} synthetic examples is less effective than \gal in \secref{sec:gal} and \secref{sec:expr}. To further verify this argument, we sample 100 instances from the synthetic RTE dataset generated by the {label-prompted GPT2, as the class-conditional LM.} Then we annotate these examples using a human annotator, {GPT2 classifier} and RoBERTa classifier. Finally, we compute the Accuracy, F1, Precision and Recall scores between {human labels and GPT2 labels}, between human labels and RoBERTa labels, and between human labels and conditioned labels used by GPT2 when data was generated. 
\tabref{tab:human_eval} shows that class-conditional LM has difficulty generating sentences retaining the semantics or pragmatics of a specified category, which also corroborates our theoretical analysis in \secref{sec:gal}. On the other hand, {discriminative models, such as GPT2 classifier and RoBERTa classifier,} are able to produce higher quality labels that correlate better with human annotations.

\subsection{Generated Unlabeled Examples Annotated with Pseudo Labels}
We provide some synthetic sentences generated by \gal in \tabref{tab:sst2} and \ref{tab:qqp}.
\begin{table*}[h!]
\caption{\textbf{SST-2}: Two labeled examples, along with 3 nearest neighbors (based on RoBERTa representations) from our synthetic dataset. We include \textbf{labels} for original examples
    and \textbf{pseudo-labels} for synthetic examples in parenthesis.}
    \centering
    \scalebox{0.9}{
    \begin{tabular}{p{0.9\linewidth}}
    \toprule
        are more deeply thought through than in most ` right-thinking ' films (\textbf{positive})\\
\midrule
KNN:\\
1: is far more sophisticated , insightful and thought-provoking than his previous films . (\textbf{positive})\\
2: is more sophisticated than its more obvious and less-than-dazzling counterparts (\textbf{positive})\\
3: is about as well-thought as the idea of a bad hair day , (\textbf{negative})\\
        \midrule\midrule
        contains no wit , only labored gags (\textbf{negative})\\
\midrule
KNN: \\
1: lacks insight , and lacks empathy (\textbf{negative})\\
2: has little humor or intelligence (\textbf{negative})\\
3: lacks all wit and humanity (\textbf{negative})\\
       \bottomrule
         
    \end{tabular}
    }

    \label{tab:sst2}
\end{table*}

\begin{table*}[h!]
\caption{\textbf{QQP}: Two labeled examples, along with 3 nearest neighbors (based on RoBERTa representations) from our synthetic dataset. We include \textbf{labels} for original examples
    and \textbf{pseudo-labels} for synthetic examples in parenthesis.}
    \centering
    \small{
    \begin{tabular}{p{0.9\linewidth}}
    \toprule
       How is the life of a math student? Could you describe your own experiences? [SEP]   Which level of prepration is enough for the exam jlpt5? (\textbf{not duplicated})\\
       \midrule
       KNN: \\

1: What are the best courses for a mechanical engineering student? [SEP] What is the best course to do after completing a B.Tech in mechanical engineering? (\textbf{not duplicated})\\
2: How much marks are needed to get through the GATE with electronics? [SEP] What is the average score of the Gate EE exam? What are the cut-offs? (\textbf{not duplicated})\\
3: What is the best time table for students to prepare for IAS? [SEP] How can one study for IAS in a best time? (\textbf{not duplicated})\\
\midrule\midrule
How does an IQ test work and what is determined from an IQ test? [SEP] How does IQ test works? (\textbf{duplicated})\\
\midrule
KNN: \\
1: What is the average IQ of the U.S. population?  [SEP] How does an IQ test work? (\textbf{not duplicated})\\
2: Is the Iq test an effective way to measure intelligence? [SEP] How do IQ tests work? (\textbf{duplicated})\\
3: How is an IQ test on a scale from 1 to 100 scored? [SEP] How do you get your IQ tested? (\textbf{not duplicated})\\
       \bottomrule
         
    \end{tabular}
    }
    \label{tab:qqp}
\vspace{1cm}
\end{table*}

\section{GAL on Image Classification Tasks}
\label{app:ic-task}

\subsection{CIFAR-10 and Fashion MNIST}

As a proof of concept, in addition to NLP tasks, we assess the effectiveness of \gal on CIFAR-10 \citep{cifar} and Fashion MNIST \citep{xiao2017fashion} as well. We adopt the
NCSN model of~\cite{song2019generative_ncsn} as the task-specific generative model.
We use the CIFAR-10 model provided by the authors and train a model on Fashion MNIST using the same configuration as CIFAR-10.
We select the model checkpoint resulting in the best FID score vs. training set~\citep{heusel2017gans} based on 1000 samples.
We then use the NCSN models to generate up to 10$\times$ synthetic unlabeled data, \ie~500K for CIFAR-10 and 600K for Fashion MNIST.
See Appendix~\ref{sec:example} for representative samples.

\begin{wraptable}{r}{7cm}
\vspace{-3mm}
\setlength\tabcolsep{3pt}
\small
\caption{Classification error rates on CIFAR-10 test set with varying amounts of synthetic data for three different model architectures. Reported results are the average of 3 independent runs.}
\begin{tabular}{lccc}
\toprule
{\textbf{Model}} &  VGG19                & ResNet110  & WRN28-10 \\
{\textbf{\# params}} & 1.74M                & 20.11M  &  36.48M  \\
\midrule
Baseline            & 6.62                  & 5.85     & 3.87 \\
\midrule
\gal  $1\times$     & 5.97                   & 5.13     &  3.75 \\
\gal  $5\times$     & 5.80                  & 5.11      &  3.25 \\
\gal  $10\times$    & \textbf{5.65} & \textbf{5.10}     &  \textbf{3.23} \\
\bottomrule
\end{tabular}
\label{tab:cv-cifar10}
\end{wraptable}

We adopt FixMatch~\citep{sohn2020fixmatch} to conduct semi-supervised learning on vision tasks, since
FixMatch has shown promising results on CIFAR-10.
Specifically, we train a classifier on mini-batches of intertwined labeled and unlabeled data (synthetic).
In each iteration, we obtain pseudo-labels for the unlabeled data, but filter unlabeled examples based on classifier's confidence, 
\ie examples are kept on which the largest class probability exceeds $\tau$.
Weak augmentation is used to define pseudo labels, but strong augmentations are used to obtain student model's predictions.
We randomly sample from the strong augmentations list defined in RandAugment~\citep{cubuk2020randaugment}.
We only apply strong augmentations to the synthetic samples and not the original labeled data 
to ensure a fair comparison with the baseline. 

We conduct experiments on three different convolutional neural network architectures: VGG19~\citep{simonyan2014very_vgg}, WideResnet28-10~\citep{zagoruyko2016_wideresnet}, and ResNet110~\citep{he2016deep_resnet}.
For the full list of hyperparameters and other implementation details, please refer to Appendix~\ref{sec:training_details}.
Each classifier is trained for 200 epochs and 3 synthetic datasets of size ($1\times$, $5\times$, $10\times$) of the training dataset are used.

Table~\ref{tab:cv-cifar10} shows that \gal achieves an average error reduction of $0.78\%$ over the baseline on CIFAR-10 across the 3 architectures tested.
Further, it appears that the larger the synthetic dataset size, the better the performance of \gal is.
We note that the reported results are the average of 3 independent runs.
\tabref{tab-cv-fashion} presents \gal results on Fashion MNIST dataset. Similar to CIFAR-10, we observe a performance improvement across the three architectures.
Our image classification experiments confirm that even when the generative model of \gal is not pretrained on open domain data and solely trained on the dataset at hand,
\gal can offer significant improvements.

\begin{table}[h!]
\caption{Classification error rates on Fashion MNIST test set with varying amounts of synthetic data for three different model architectures. Results reported are the average over 3 independent runs.}
\small
\centering
\begin{tabular}{lccc}
\toprule
\multicolumn{1}{c}{\textbf{Model}}   & VGG19                & WRN28-2              & ResNet110            \\
\multicolumn{1}{c}{\textbf{\# params}} & 1.74M                & 1.98M                & 20.11M               \\ \midrule
Baseline           & 5.41     & 4.92                & 5.21                \\
\midrule
\gal~$1\times$       & 5.06          & \textbf{4.63}        & \textbf{4.74} \\
\gal~$5\times$       & 5.14          & 4.85                 & 4.85  \\
\gal~$10\times$      & \textbf{4.90} & 4.74                 & 4.75  \\          
\bottomrule
\end{tabular}
\label{tab-cv-fashion}
\end{table}

\newpage 
\subsection{Generated Unlabeled Examples Annotated with Pseudo Labels}
\label{sec:example}

\begin{figure*}[h!]
 \centering
 \begin{tabular}{@{}r@{\hspace{.2cm}}c@{}}
 \raisebox{.85cm}{truck} & \includegraphics[width=0.88\textwidth]{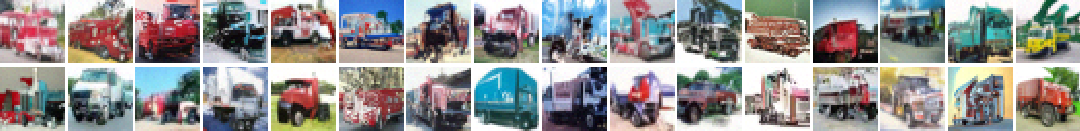} \\
 \raisebox{.85cm}{ship} & \includegraphics[width=0.88\textwidth]{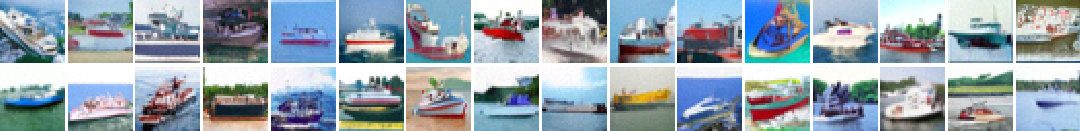} \\
 \raisebox{.85cm}{horse} & \includegraphics[width=0.88\textwidth]{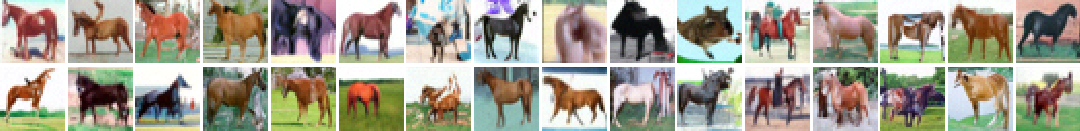} \\
 \raisebox{.85cm}{frog} & \includegraphics[width=0.88\textwidth]{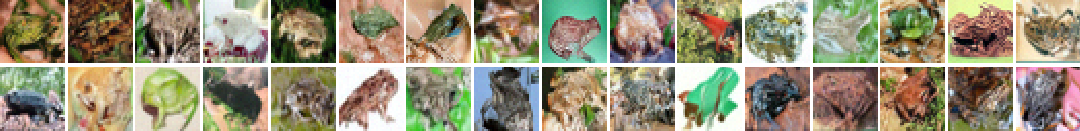} \\
 \raisebox{.85cm}{dog} & \includegraphics[width=0.88\textwidth]{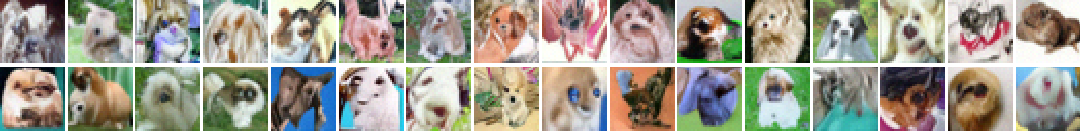} \\
 \raisebox{.85cm}{deer} & \includegraphics[width=0.88\textwidth]{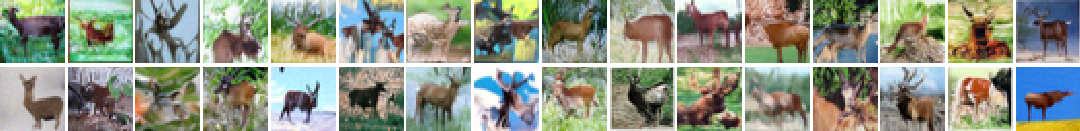} \\
 \raisebox{.85cm}{cat} & \includegraphics[width=0.88\textwidth]{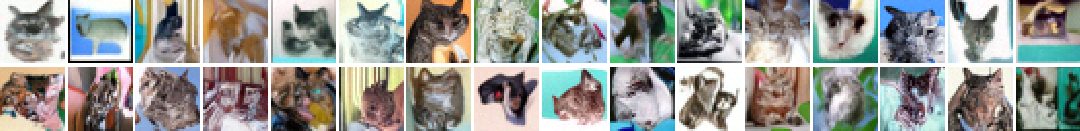} \\
 \raisebox{.85cm}{bird} & \includegraphics[width=0.88\textwidth]{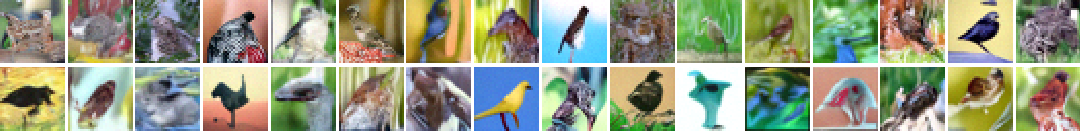} \\
 \raisebox{.85cm}{car} & \includegraphics[width=0.88\textwidth]{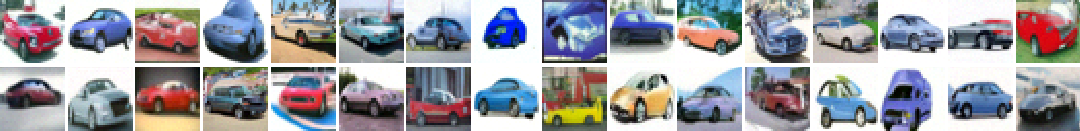} \\
 \raisebox{.85cm}{airplane} & \includegraphics[width=0.88\textwidth]{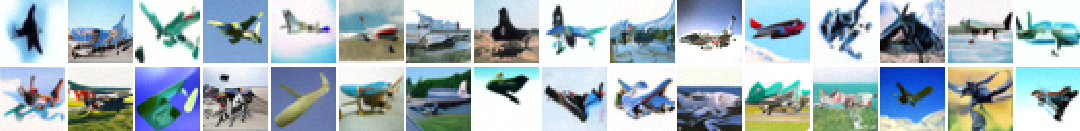}
\end{tabular}
\caption{CIFAR-10 synthetic samples generated by NCSN~\citep{song2019generative_ncsn} and corresponding pseudo-labels. Images are filtered based on a confidence threshold of $\tau = 0.95$ and categorized based on pseudo-labels. For each category, 16 random samples are shown.}
\label{fig:synth_examples_cifar10}
\end{figure*}

\begin{figure*}[h!]
 \centering
 \begin{tabular}{@{}r@{\hspace{.2cm}}c@{}}
 \raisebox{.3cm}{boot} & \includegraphics[trim={0 2.28cm 0 0},clip,width=0.88\textwidth]{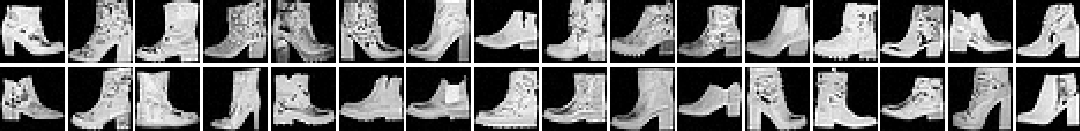} \\
 \raisebox{.3cm}{bag} & \includegraphics[trim={0 2.28cm 0 0},clip,width=0.88\textwidth]{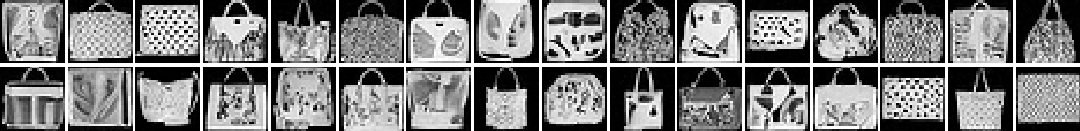} \\
 \raisebox{.3cm}{sneaker} & \includegraphics[trim={0 2.28cm 0 0},clip,width=0.88\textwidth]{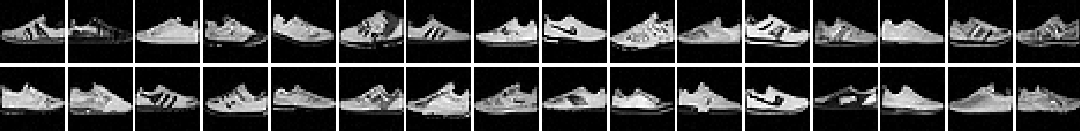} \\
 \raisebox{.3cm}{shirt} & \includegraphics[trim={0 2.28cm 0 0},clip,width=0.88\textwidth]{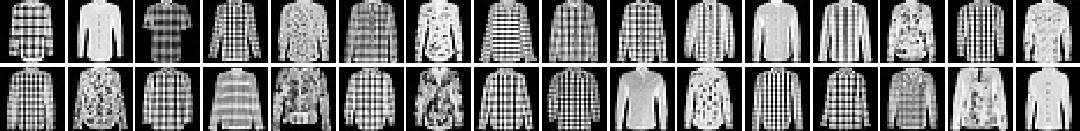} \\
 \raisebox{.3cm}{sandal} & \includegraphics[trim={0 2.28cm 0 0},clip,width=0.88\textwidth]{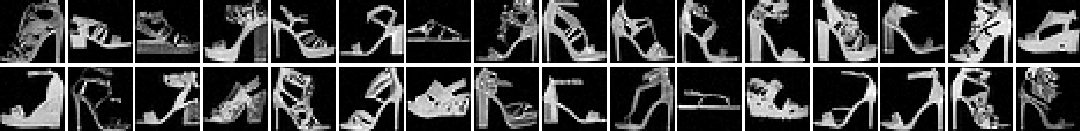} \\
 \raisebox{.3cm}{coat} & \includegraphics[trim={0 2.28cm 0 0},clip,width=0.88\textwidth]{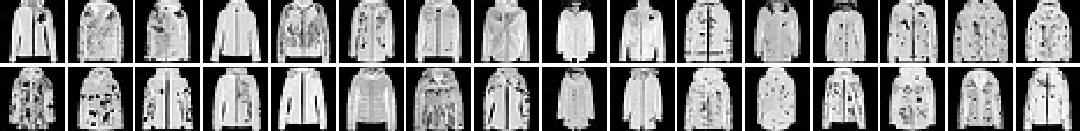} \\
 \raisebox{.3cm}{dress} & \includegraphics[trim={0 2.28cm 0 0},clip,width=0.88\textwidth]{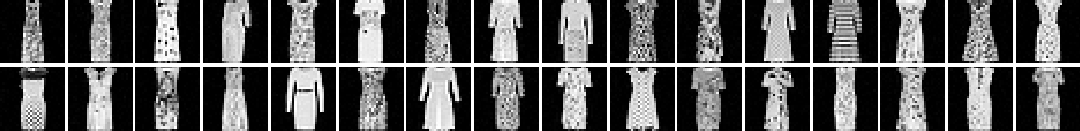} \\
 \raisebox{.3cm}{pullover} & \includegraphics[trim={0 2.28cm 0 0},clip,width=0.88\textwidth]{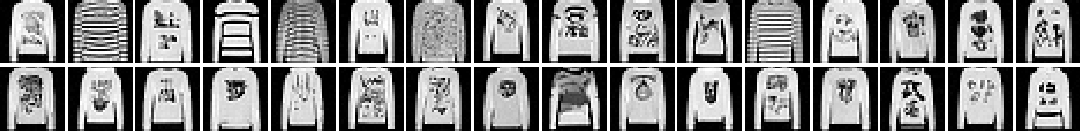} \\
 \raisebox{.3cm}{trouser} & \includegraphics[trim={0 2.28cm 0 0},clip,width=0.88\textwidth]{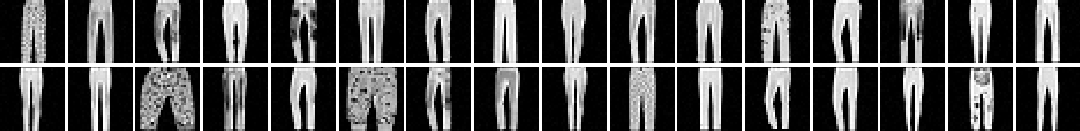} \\
 \raisebox{.3cm}{t-shirt} & \includegraphics[trim={0 2.28cm 0 0},clip,width=0.88\textwidth]{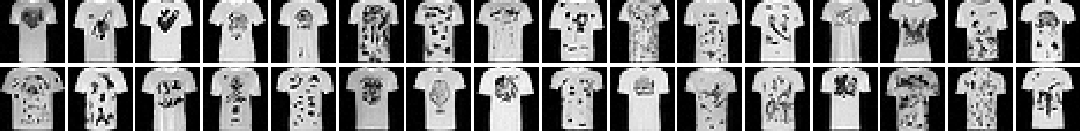}
\end{tabular}
\caption{Fashion MNIST synthetic samples generated by NCSN~\citep{song2019generative_ncsn} and pseudo-labels. Images are filtered based on a confidence threshold of $\tau = 0.95$ and categorized based on pseudo-labels. For each category, 16 random samples are shown.}
\label{fig:synth_examples_fmnist}
\end{figure*}

\vfill

\section{Training Details}
\label{sec:training_details}
We use the fairseq codebase~\citep{ott2019fairseq} for the self-training over GLUE benchmark. Training details are summarized in \tabref{tab:training}. We use the HuggingFace codebase~\citep{wolf-etal-2020-transformers} for KD experiments. All KD experiments are trained for 5 epochs with a learning rate of 2e-5 and a batch size of 32. All experiments are run on a single Nvidia V100 GPU.

\begin{table*}[h!]
    \caption{Training details for self-training over GLUE benchmark.}
    \centering
    \begin{tabular}{lcccccccccc}
    \toprule
            & \textbf{MNLI}	& \textbf{CoLA}	& \textbf{SST-2} & \textbf{MRPC} & \textbf{STS-B} & \textbf{QQP} & \textbf{QNLI} & \textbf{RTE}  \\
    \midrule 
    lr & 1e-5 & 1e-5 & 1e-5	 & 1e-5	 & 2e-5	 & 1e-5	 & 1e-5	 & 2e-5 \\
    \#sent. &32	 & 16 & 32 & 16 & 16 & 32 & 32 & 16  \\
    warmup steps & 7432 & 320 & 1256 & 137 & 214 & 28318 & 1986 & 122 \\
    validate steps &12386 & 535 & 2093 & 203 & 360 & 11307 & 3310 & 203\\
    \#epoch &2 & 2 & 2 & 2 & 2 & 2 & 2 & 2 \\

     \bottomrule
    \end{tabular}
    \vspace*{-.1cm}
    \vspace*{-.1cm}
    \label{tab:training}
\end{table*}


   


\begin{table*}[h!]
\caption{Training details for CV experiments}
\centering
\scalebox{0.85}{
\begin{tabular}{llllc}
\hline
\multicolumn{1}{c}{\textbf{Parameter}} &
   &
  \multicolumn{1}{c}{\textbf{Description}} &
   &
  \textbf{Value} \\ 
 \hline
$\tau$           &  & Pseudo-labeling confidence threshold        &  & 0.95    \\
batch size         &  & Number of labeled images per batch                                 &  & 64      \\
$\mu$               &  & Ratio between number of unlabeled and labeled images in each batch &  & 7        \\
images per epoch &  & Number of labeled images per epoch                                 &  & 65536  \\
\#epoch        &  & Number of epochs of training                                       &  & 200      \\
lr      &  & learning rate max value (10 epochs warmup then cosine decay)       &  & 0.03     \\
weight decay      &  & Weight decay regualrization coefficient                            &  & $5.00\times10^{-4}$ \\
momentum           &  & Nesterov momentum for SGD optimizer                                &  & 0.90      \\
\hline
\end{tabular}
}
\label{tab:training_cv}
\end{table*}

For the CV tasks, we first use the official implementation of NCSN ~\citep{song2019generative_ncsn} to generate the synthetic images for CIFAR-10 and Fashion MNIST. We use the pretrained checkpoints provided by the authors for the generation of synthetic CIFAR-10 images and we train a new generative model for Fashion MNIST from scratch with the same hyperparameters of the CIFAR-10 network. After generating the synthetic images, we apply \gal using a FixMatch-like setup~\citep{sohn2020fixmatch}, using the 
hyperparameters listed in \tabref{tab:training_cv}. We follow ~\citet{cubuk2020randaugment} for strong augmentations. Finally, the backbone of the classifiers is from this codebase: \url{https://github.com/bearpaw/pytorch-classification}.

\fi

\end{document}